\documentclass[journal]{IEEEtran}
\newcommand{\textwidthh}{0.48\textwidth}
\newcommand{\textwidthhh}{0.24\textwidth}

\usepackage{amsmath,amssymb}
\usepackage{lscape}

\newcommand{\argmin}{\mathop{\mathrm{argmin}}}

\usepackage{subfigure}
\usepackage[dvipdfmx]{graphicx}
\usepackage{algorithm,algorithmic}
\usepackage{color}
\usepackage{url}

\newcommand{\bm}{\boldsymbol}
\newcommand{\ten}[1]{ \boldsymbol{\mathcal #1}}
\newcommand{\bbR}[1]{\mathbb{R}^{#1}}
\newcommand{\jump}[1]{\ensuremath{ [\![#1]\!] } }

\begin{document}

\title{Smooth PARAFAC Decomposition for Tensor Completion}

\author{Tatsuya Yokota, ~\IEEEmembership{Member,~IEEE,}
        Qibin Zhao, ~\IEEEmembership{Member,~IEEE,}
        and Andrzej Cichocki, ~\IEEEmembership{Fellow,~IEEE}}

\markboth{Yokota \MakeLowercase{\textit{et al.}}: Smooth PARAFAC decomposition for Tensor Completion (Version 3)}
{Yokota \MakeLowercase{\textit{et al.}}: Smooth PARAFAC decomposition for Tensor Completion (Version 3)}

\IEEEcompsoctitleabstractindextext{
\begin{abstract}
In recent years, low-rank based tensor completion, which is a higher-order extension of matrix completion, has received considerable attention.
However, the low-rank assumption is not sufficient for the recovery of visual data, such as color and 3D images, where the ratio of missing data is extremely high.
In this paper, we consider ``smoothness'' constraints as well as low-rank approximations, and propose an efficient algorithm for performing tensor completion that is particularly powerful regarding visual data.
The proposed method admits significant advantages, owing to the integration of smooth PARAFAC decomposition for incomplete tensors and the efficient selection of models in order to minimize the tensor rank. Thus, our proposed method is termed as ``smooth PARAFAC tensor completion (SPC).''
In order to impose the smoothness constraints, we employ two strategies, total variation (SPC-TV) and quadratic variation (SPC-QV), and invoke the corresponding algorithms for model learning.
Extensive experimental evaluations on both synthetic and real-world visual data illustrate the significant improvements of our method, in terms of both prediction performance and efficiency, compared with many state-of-the-art tensor completion methods. 
\end{abstract}
\begin{keywords}
Tensor completion for images, smoothness, low-rank tensor approximation, CP model, PARAFAC model, total variation (TV), quadratic variation.
\end{keywords}}

\maketitle

\IEEEdisplaynotcompsoctitleabstractindextext

\IEEEpeerreviewmaketitle

\section{Introduction}
Completion is a procedure that facilitates the estimation of the values of missing elements of array data, using only the available elements and structural properties of the data.
Clearly, if there is no relationship between the missing elements and the available elements, completion is not possible.
However, real world data usually exhibits some correlations, latent factors, symmetry, continuity, or repetition, in which case completion is often possible.
For example, when a vector consists of the sampled values of some continuous function that has several missing values, some interpolation methods can be used for completion, such as simple linear interpolation, spline interpolation, and polynomial interpolation.
When a given matrix has several missing entries and the low-rank matrix factorization exists, then a low-rank structure can be used for completion, by approximating the given matrix using the low-rank factorization model.
Such completion techniques are closely associated with computer vision, pattern recognition, and compressed sensing \cite{criminisi2004region,geng2011face,li2013compressed}.

Techniques for vector/matrix completion have been comprehensively researched, and many sophisticated methods exist.
Furthermore, techniques for ``tensor'' completion have attracted attention in recent years, because of their potential applications and flexibility.
A tensor is a multi-dimensional array, vectors and matrices can be considered as first-order and second-order tensors, respectively.
For example, the data for a color-image is a third-order tensor, because it consists of three color shading images of red, green, and blue.
Similarly, the data for a color video is a fourth-order tensor, because it consists of multiple frames of color images.
There have been several papers that have attempted the completion color images using matrix completion \cite{ji2010robust,han2014linear,li2014non}.
In \cite{han2014linear}, the authors proposed to complete each color shading image separately, and subsequently concatenate them.
However, such an approach ignores the natural multi-dimensional structure of tensors, and thus neglects some important information.
In \cite{ji2010robust,li2014non}, the authors applied a matrix completion technique to matrices that consist of set of similar patches selected by some patch matching algorithms. However, patch matching is generally time consuming, and it does not work well when there is an extremely high ratio of missing data.

On the other hand, tensor completion methods have shown significant progress in the completion of color images by exploiting the structural information of 3rd-order tensors.
The state-of-the-art methods for tensor completion consist of approaches of two types.
The first approach, called nuclear norm minimization, is based on the low-rank property \cite{liu2009tensor,gandy2011tensor,liu2013tensor}.
The nuclear norm of a matrix is defined as the sum of all singular values, and is a convex envelope for the rank of matrix \cite{recht2010guaranteed}.
The nuclear norm of the tensor was first introduced and applied to tensor completion in \cite{liu2009tensor}.
Subsequently, some improved algorithms were proposed in \cite{gandy2011tensor,liu2013tensor}.
The minimization of tensor nuclear norms has also been applied in various contexts, such as error correction \cite{li2010optimum,li2010tensor}, medical imaging \cite{lv2011hole}, compression \cite{li2010tensor_comp}, and saliency detection \cite{yan2010visual}.
The second approach involves the use of low-rank tensor decomposition techniques, which have been proposed in \cite{acar2011scalable,sorber2013optimization,kressner2014low,tan2014tensor,chen2014simultaneous,rai2014scalable,zhao2015bayesian}.
Tensor factorization is a method for decomposing an $N$th-order tensor into another $N$th-order tensor of smaller size, termed as the ``core tensor,'' and $N$ factor matrices.
There are two factorization models: Tucker decomposition \cite{Tuck1963a} and polyadic decomposition \cite{hitchcock1927expression}.
The polyadic decomposition is also called several different names such as CANDECOMP (canonical decomposition) \cite{carroll1970analysis}, PARAFAC (parallel factor) decomposition \cite{harshman1970foundations}, and CP (CANDECOMP/PARAFAC) decomposition \cite{kiers2000towards}.
In the present paper, we usually refer to this model as PARAFAC/polyadic decomposition (PD) model.
A core tensor of PD takes the form of a super-diagonal tensor, and a core tensor of Tucker decomposition takes the form of a general tensor.
Thus, PD constitutes a special case of Tucker decomposition.
Similarly to the singular value decomposition (SVD) of matrix factorization, the minimum size of a core tensor corresponds to the rank of the original tensor.
In \cite{chen2014simultaneous}, the authors consider the low rank Tucker decomposition model for tensor completion and the minimization of the nuclear norms of individual factor matrices.
In \cite{zhao2015bayesian}, the authors consider the low-rank PD model for tensor completion, by using a Bayesian framework to solve tuning parameter problems.
Furthermore, the authors of \cite{chen2014simultaneous} and \cite{zhao2015bayesian} introduced a factor prior regularization, in order to yield improved results for visual data.

In this paper, we investigate smoothness constraints for PD model based tensor completion.
Smoothness is an important property that is embedded in many examples of real world data, such as natural images/videos, some spectral signals, and bio-medical data.
For example, low-frequency (smooth) components of natural images computed by discrete cosine transform have higher magnitude than high-frequency components.
This is a reason why the JPEG encoder is efficient and widely used in practice, and it indicates the usability of smoothness property for natural image processing.
When we assume original images and additive noise are respectively smooth and non-smooth, a smoothness constraint helps to preserve a smooth original image and to remove non-smooth noise.
In fact, matrix/tensor factorizations with smoothness constraints have many applications in which they are robust in the presence of noisy signals, such as blind source separation \cite{zdunek2008blind,zdunek2012approximation,zdunek2014b}, video structuring \cite{essid2013smooth}, visual parts extraction \cite{yokota2015smooth}, genomic data analysis \cite{witten2009penalized}, and brain signal analysis \cite{chen2006constrained}.

Total variation (TV) \cite{rudin1992nonlinear}, which is defined as the $l_1$-norm of the difference of neighbor elements, is often used to impose piece-wise smoothness constraints.
Many methods have applied the TV approach in image restoration and denoising \cite{osher2005iterative,goldfarb2005second,wang2008new,beck2009fast,ng2010solving,chan2011augmented}.
Furthermore, the TV constraint has been already applied to matrix completion in \cite{dai2009physics,han2014linear}, but until now it has not been investigated for tensors.
Therefore, we apply the TV constraint to PD model based tensor completion.
Furthermore, we investigate an additional smoothness constraint, which is defined as the ``$l_2$-norm'' of the difference between neighbor elements, and is a stronger constraint than the TV constraint.
Our concept of smoothness is not same as in other existing smoothness based completion methods \cite{rudin1992nonlinear,han2014linear,chen2014simultaneous,zhao2015bayesian}.
The key point of our approach is to assume that ``latent component vectors'' of factor matrices are smooth, and different levels of smoothness are enforced on different components adaptively.
The details of the novelty of our model are discussed in Section~\ref{sec:novelty}.

In addition, we propose a new strategy for a rank determination algorithm for the smooth PD model.
A general strategy of the above mentioned ``state-of-the-arts'' algorithms involves setting the upper-bound of the rank for a tensor and optimizing the rank by using some procedures to remove redundant components, such as singular value shrinkage.
However, if we impose our smoothness constraint on a tensor, the number of required components increases, and we are not able to determine an upper-bound.
Therefore, we propose a reverse strategy, estimating the rank of the tensor (i.e., the optimal number of components) by increasing the rank step by step, starting with a rank-one tensor. Our algorithm will stop when the smooth PD model fits the observed data sufficiently well.
We carried out experiments involving tensor completion problems by using several difficult benchmarks of color images with different types of missing pixels (e.g., text masked, scratched images, and random pixels missing), incomplete MRI data, and multi-way structured facial data.

The remainder of this paper is organized as follows.
In Section \ref{sec:notations}, we explain the notations used in this paper.
Section \ref{sec:existing} reviews several existing matrix/tensor completion methods.
In Section \ref{sec:proposed}, we propose novel algorithms for tensor completion based on the smooth PD model.
In Section \ref{sec:experiments}, we investigate the performance and applications of our algorithms, and compare them with some state-of-the-art methods.
In Section \ref{sec:disc}, we discuss several aspects of our work.
Finally, we present our conclusions in Section \ref{sec:conclusions}.

\subsection{Preliminaries and Notations} \label{sec:notations}
A vector is denoted by a bold small letter $\bm a \in \bbR{I}$.
A matrix is denoted by a bold capital letter $\bm A \in \bbR{I \times J}$.
A higher-order ($N \geq 3$) tensor is denoted by a bold calligraphic letter $\ten{A} \in \bbR{I_1 \times I_2 \times \cdots \times I_N}$.
The $i$th entry of a vector $\bm a \in \bbR{I}$ is denoted by $\bm a(i)$, and the $(i,j)$th entry of a matrix $\bm A \in \bbR{I \times J}$ is denoted by $\bm A(i,j)$.
The $(i_1, i_2, ..., i_N)$th entry of an $N$th-order tensor $\ten{A}$ is denoted by $\ten{A}_{i_1 i_2 \cdots i_N}$, where $i_n \in \{1, 2, ..., I_n\}$ and $n \in \{1, 2, ..., N\}$.
The Frobenius norm of an $N$th-order tensor is defined by $|| \ten{X} ||_F := \sqrt{\sum_{i_1, i_2, ..., i_N} \ten{X}_{i_1i_2 \cdots i_N}^2}$.

A mode-$k$ unfolding of a tensor $\ten{X}$ is denoted as $\bm X_{(k)} \in \bbR{I_k \times \Pi_{n \neq k} I_n }$.
For example, a first mode unfolding of a third-order tensor $\ten{X} \in \bbR{I_1 \times I_2 \times I_3}$ is given by $\bm X_{(1)}(i_1,(i_3-1)I_2 + i_2) = \ten{X}_{i_1i_2i_3}$, where $i_n \in \{1, 2, ..., I_n \}$ and $n \in \{1, 2, 3\}$.
A mode-$k$ multiplication between a tensor $\ten{X} \in \bbR{I_1 \times I_2 \times \cdots \times I_N}$ and a matrix/vector $\bm A \in \bbR{I_k \times R}$ is denoted by $\ten{Y} = \ten{X} \times_k \bm A^T \in \bbR{I_1 \times \cdots \times I_{k-1} \times R \times I_{k+1} \times \cdots \times I_N}$, with entries given by $\ten{Y}_{i_1 \cdots i_{k-1} r i_{k+1} \cdots i_N} = \sum_{i_k} \ten{X}_{i_1 \cdots i_{k-1} i_k i_{k+1} \cdots i_N} \bm A(i_k,r) $, and we have $\bm Y_{(k)} = \bm A^T \bm X_{(k)}$.

\section{Brief Review of Existing Methods}\label{sec:existing}
In this section, we explain the methodology for matrix/tensor completion based on the low-rank assumption, and review the several state-of-the-art methods.

Concerning matrix data, the most popular and basic approach for matrix completion is the minimization of the matrix rank, subject to fitting the available elements:
\begin{align}
\mathop{\text{minimize}}_{\bm X} \ \ \text{rank}(\bm X),\ \  \text{s.t. } \bm X_\Omega = \bm T_\Omega,
\end{align}
where $\bm X$ is a completed output matrix, $\bm T$ is an incomplete input matrix, $\Omega$ denotes the indices of the available elements of $\bm T$, the equation $\bm X_\Omega = \bm T_\Omega$ means that $\bm X(i,j) = \bm T(i,j),\ \forall (i,j) \in \Omega$, and $\text{rank}(\bm X)$ denotes the rank of matrix $\bm X$.
Such a rank minimization approach has been applied in various research fields, such as machine learning \cite{amit2007uncovering,evgeniou2007multi} and bioinformatics \cite{troyanskaya2001missing}.
If an original matrix is structured as a low-rank matrix, then this rank minimization approach can be used to obtain an estimation of the ground truth matrix.
However, rank is not convex function with respect to $\bm X$, and rank minimization is generally NP-hard \cite{gillis2011low}.
Therefore, the nuclear-norm minimization method is widely used in practice:
\begin{align}
\mathop{\text{minimize}}_{\bm X} \ \   || \bm X ||_*,\ \  \text{s.t. } \bm X_\Omega = \bm T_\Omega, \label{eq:matrix_completion}
\end{align}
where $|| \bm X ||_* = \sum_i \sigma_i(\bm X)$ denotes the nuclear norm, and $\sigma_i(\bm X)$ denotes the $i$th largest singular value of the matrix $\bm X$.
In order to solve the optimization problem \eqref{eq:matrix_completion}, the following augmented Lagrange function can be applied:
\begin{align}
  L_A (\bm X,& \bm Y, \bm W, \beta) \notag\\ &= || \bm X ||_* - \langle \bm W, \bm X - \bm Y \rangle + \frac{\beta}{2}||\bm X - \bm Y||_F^2, \label{eq:Lagrange_ADMM}
\end{align}
where $\bm Y_\Omega = \bm T_\Omega$ is an auxiliary matrix, $\bm W$ is a Lagrangian multiplier matrix, $\beta$ is a penalty (trade-off) parameter for the augmented term $||\bm X - \bm Y||_F^2$, and $\beta$ is increased in each iteration step.
This problem can be optimized by applying the alternating direction method of multipliers (ADMM) \cite{bertsekas1989parallel}.
There are several existing papers that have applied the ADMM scheme to matrix or tensor completion problems \cite{lin2010augmented,gandy2011tensor,chen2012matrix,liu2013tensor}.

The smoothness property of data is often assumed for image completion.
If we consider $\bm X$ to be a gray-scale image, and individual entries stand for the values of individual pixels, then the difference between neighbor pixels should typically be small.
In \cite{rudin1992nonlinear}, the minimization of the total variation (TV) was employed as a smoothness constraint, which was defined by
\begin{align}
  || \bm X ||_{TV} := \sum_{i,j} \sqrt{ X_v(i,j)^2 + X_h(i,j)^2 },
\end{align}
where $X_v(i,j) := \bm X(i+1,j) - \bm X(i,j)$ and $X_h(i,j) := \bm X(i,j+1) - \bm X(i,j)$.
Image completion via TV minimization was proposed in \cite{dai2009physics}, and formulated in the form of the following optimization problem:
\begin{align}
  \mathop{\text{minimize}}_{\bm X} \ \ || \bm X ||_{TV},\ \ \text{s.t. } \bm X_\Omega = \bm T_\Omega. \notag
\end{align}
This problem is convex, and can be solved using gradient descent methods \cite{dai2009physics}.

Furthermore, a combination of low-rank approximation and smoothness constraints was proposed in \cite{han2014linear}.
Usually, natural images yield both of these structural features, and this combination is quite efficient.
The most straightforward approach for the combination of low-rank approximation and smoothness constraints is implemented by the following optimization problem:
\begin{align}
\mathop{\text{minimize}}_{\bm X} \ \ || \bm X ||_* + \gamma || \bm X ||_{TV},\ \ \text{s.t. } \bm X_\Omega = \bm T_\Omega,
\end{align}
where $\gamma$ is a trade-off parameter between nuclear norm minimization and TV minimization.
In \cite{han2014linear}, a modified linear total variation was defined, as
\begin{align}
  \text{LTV}(\bm X) := \sum_{i,j} \{ X_v(i,j)^2 + X_h(i,j)^2 \} ,
\end{align}
and a smooth low-rank matrix completion method was proposed in the form of
\begin{align}
  \mathop{\text{minimize}}_{\bm X} \ \  || \bm X ||_* + \gamma \text{LTV}(\bm X),\ \  \text{s.t. } \bm X_\Omega = \bm T_\Omega. \label{eq:LTVNN}
\end{align}
The optimization problem \eqref{eq:LTVNN} was referred to as the linear total variation approximate regularized nuclear norm (LTVNN) minimization problem, and this problem can be solved using an ADMM like optimization scheme.
This approach can be considered as the state-of-the-art method for low-rank image completion. However, LTVNN may be not useful for tensor completion.
Concerning color images (third-order tensors), the authors separated an image into red, green, and blue frames, and applied the completion method to each color frame separately.
However, when individual color frames are quite similar, such a separated method may not be effective.

Tensor completion is a natural extension of matrix completion with respect to the data structure, and it can use such structural information more effectively than matrix completion.
A basic method for tensor completion was proposed using simple low-rank tensor completion (SiLRTC) in \cite{liu2013tensor}, and this was then formulated as the following constrained optimization problem:
\begin{align}
\mathop{\text{minimize}}_{\ten{X}, \bm Y^{(1)},... ,\bm Y^{(N)}}\ \ & \sum_{i=1}^N \left\{ \alpha_i || \bm Y^{(i)} ||_* + \frac{\beta}{2}||\bm X_{(i)} - \bm Y^{(i)}||^2_F \right\} ,\ \\ \notag \text{s.t. } & \ \ \ten{X}_\Omega = \ten{T}_\Omega, \label{eq:SiLRTC}
\end{align}
where $\ten{X}$ is a completed output tensor, $\ten{T}$ is an incomplete input tensor, $\bm Y^{(i)}$ is a low-rank matrix corresponding to the mode-$i$ matricization form $\bm X_{(i)}$, and $\alpha_i$ and $\beta$ are weight parameters for individual cost functions.
Essentially, this method attempts to minimize the tensor nuclear norm, defined as $\sum_{i} \alpha_i || \bm X_{(i)} ||_*$, which is a generalization of the matrix nuclear norm.
Instead of the minimization of $\sum_{i} \alpha_i || \bm X_{(i)} ||_*$, we minimize the nuclear norm of alternate parameters $\bm Y^{(i)}$, the mean squared error between alternate parameters $\bm Y^{(i)}$, and the mode-$i$ unfolding of $\ten{X}$.

Note that when we strictly consider the minimization of $\sum_{i} \alpha_i || \bm X_{(i)} ||_*$, constraints $\bm X_{(i)} = \bm Y^{(i)}$ should be added.
The high accuracy low rank tensor completion (HaLRTC) method \cite{liu2013tensor} was proposed on the basis of this concept, by formulating the following optimization problem:
\begin{align}
\mathop{\text{minimize}}_{\ten{X}, \ten{Y}^{(1)},... ,\ten{Y}^{(N)}} & \ \ \sum_{i=1}^N \alpha_i || \bm Y^{(i)} ||_*, \\ \notag \text{s.t. } & \ \ \ten{X} = \ten{Y}^{(i)},\  \ten{X}_\Omega = \ten{T}_\Omega,
\end{align}
where $\ten{Y}^{(i)}$ is the tensor form of matrix $\bm Y^{(i)}$.
The authors considered an augmented Lagrange function, which is an extension of \eqref{eq:Lagrange_ADMM}, and can be minimized using the ADMM optimization scheme in similar manner.

Recently, an alternative and very efficient completion method has been proposed, called simultaneous tensor decomposition and completion (STDC) \cite{chen2014simultaneous}.
This method is based on Tucker decomposition, and minimizes the nuclear norm of individual factor matrices by solving the following optimization problem:
\begin{align}
\mathop{\text{minimize}}_{\ten{G}, \bm U^{(1)}, ..., \bm U^{(N)}} &\ \  \sum_{i=1}^N \alpha_i ||\bm U^{(i)}||_* + \delta \text{tr}( \bm \Phi  \bm L \bm \Phi^T ) + \gamma ||\ten{G}||_F^2, \label{eq:opt_problem_stdc}\\ \notag
 \text{s.t. } & \ \ \ten{X} = \ten{G} \times \{ \bm U^{(i)} \}_{i=1}^N,\ \ten{X}_\Omega = \ten{T}_\Omega, 
\end{align}
where $\ten{G}$ is a core tensor, $\bm U^{(i)}$ is a factor matrix, $\ten{G} \times \{ \bm U^{(i)} \}_{i=1}^N := \ten{G} \times_1 \bm U^{(1)} \times_2 \bm U^{(2)} \times_3 ... \times_N  \bm U^{(N)}$ is the Tucker decomposition model of tensor, $\bm \Phi := (\bm U^{(1)} \otimes \cdots \otimes \bm U^{(N)})$ is a unified factor matrix that consists of all factor matrices, $\otimes$ is the Kronecker product, and $\bm L$ is a matrix designed by some prior information.
The STDC method is based on the ``low-rank Tucker decomposition'' and ``factor prior.''
The minimization of $\delta\text{tr}( \bm \Phi  \bm L \bm \Phi^T )$ can be interpreted as enforcing the similarity between individual components, which plays the role of a regularization, and provides some smoothed results for the visual data.
Thus, STDC can be considered as a tensor extension of low-rank and smooth matrix completion.
The authors considered an augmented Lagrange function, in order to solve \eqref{eq:opt_problem_stdc} in similar way.

Finally, an additional efficient and promising method for tensor completion is given by fully Bayesian CANDECOMP/PARAFAC tensor completion with mixture prior (FBCP-MP) \cite{zhao2015bayesian}.
The FBCP-MP method is based on the ``probabilistic low-rank PD'' and ``mixture prior.''
By using mixture prior, this probabilistic model enforces a kind of similarity between individual component vectors, similarly to STDC.
Furthermore, FBCP-MP finds a PD with an appropriate tensor rank by using Bayesian inference, without need to tune or adjust any parameters.

Hence, we can categorize the state-of-the art tensor completion methods into three types: low-rank-nuclear-norm, low-rank-Tucker-decomposition, and low-rank-PD with optional smoothness constraints.
The above mentioned papers on tensor completion did not discuss smoothness constraints in detail. However, smoothness is quite an important factor for visual data completion, which is obvious in matrix completion problems \cite{dai2009physics,ram2013image,han2014linear}.

\section{Proposed New Method}\label{sec:proposed}
In this section, we propose a new algorithm for PARAFAC decomposition (PD) based tensor completion with smoothness constraints.
First, the unconstrained PD model can be formulated as
\begin{align}
  \ten{Z} = \sum_{r=1}^R g_r \bm u_r^{(1)} \circ \bm u_r^{(2)} \circ \dots \circ \bm u_r^{(N)}, \label{eq:CP_decomposition}
\end{align}
where $\circ$ stands for outer product of vectors, $\bm u_r^{(n)}$ are feature vectors, called ``components'', and $g_r$ are scaling multipliers.
By denoting factor matrices by $\bm U^{(n)} = [\bm u_1^{(n)}, \bm u_2^{(n)}, ..., \bm u_R^{(n)}]$ and a super-diagonal core tensor by $\ten{G}$, such that each super-diagonal element is $\ten{G}_{rr...r} = g_r$, the PD model can be also denoted by $\ten{Z} = \jump{ \ten{G}; \bm U^{(1)}, \bm U^{(2)}, ..., \bm U^{(N)} }$.
The minimum number of components $R$ needed to satisfy equation \eqref{eq:CP_decomposition} is called the ``tensor rank'' of $\ten{Z}$.
The exact PD with a minimal $R$ is usually refereed to as canonical polyadic decomposition (CPD) or rank decomposition \cite{kolda2009tensor}.
In general, when we impose the smoothness constraint on the feature vectors $\bm u_r^{(n)}$, the minimum number of components $R$ increases, because the smoothness constraint decreases the flexibility of the decomposition model.
For this reason, it is difficult to determine the upper bound of the tensor rank of the original tensor when the smoothness constraint is imposed.
In the STDC \cite{chen2014simultaneous} and FBCP-MP \cite{zhao2015bayesian} approaches, the upper bound of the Tucker/tensor rank is first determined as some large value, and the subsequent Tucker/tensor rank is then estimated by decreasing the number of components in each iteration.
In this paper, we propose a completely different approach, which does not require the determination of the upper bound of the tensor rank, because in the proposed method we increase the number of components $R$ gradually from $1$ up to its optimal value.
We call the new method ``smooth PARAFAC tensor completion'' (SPC).

\subsection{Fundamental Problem for the Fixed Rank SPC (FR-SPC)}
In this section, we consider the fixed rank version of SPC (FR-SPC).
The optimization problem for FR-SPC is formulated as
\begin{align}
  \mathop{\text{minimize}}_{\ten{G},\bm U^{(1)},...,\bm U^{(N)}} \ \ &\frac{1}{2}|| \ten{X} - \ten{Z} ||_F^2 \label{criterion:SmCPTC}\\
 & + \sum_{r=1}^R \frac{g_r^2}{2} \sum_{n=1}^N \rho^{(n)} || \bm L^{(n)} \bm u_r^{(n)} ||_p^p,  \notag \\
                   \text{s.t. }&\ten{Z} = \sum_{r=1}^R g_r \bm u_r^{(1)} \circ \bm u_r^{(2)} \circ \dots \circ \bm u_r^{(N)}, \notag\\
 & \ten{X}_\Omega = \ten{T}_\Omega,\ \ten{X}_{\bar{\Omega}} = \ten{Z}_{\bar{\Omega}}, ||\bm u_r^{(n)}||_2 = 1,\notag \\
                               &\forall r \in\{1,...,R\}, \forall n \in \{1,...,N\} , \notag 
\end{align}
where $\ten{X}$ is a completed output tensor in which the missing entries are filled using smooth PD approximation $\ten{Z}$.
Thus, the constraints $\ten{X}_\Omega = \ten{T}_\Omega$ and $\ten{X}_{\bar{\Omega}} = \ten{Z}_{\bar{\Omega}}$ stand for
\begin{align}
 \ten{X}_{i_1i_2...i_N} = \left\{ \begin{array}{ll} \ten{T}_{i_1i_2...i_N} & (i_1, i_2, ..., i_N) \in \Omega \\ \ten{Z}_{i_1i_2...i_N} & \text{otherwise} \end{array} \right. .
\end{align}
Next, $\bm \rho = [\rho^{(1)}, \rho^{(2)}, ..., \rho^{(N)}]^T$ is a smoothness parameter vector, $p \in \{1,2\}$ is a parameter for selecting the types of smooth constraints, and the matrix $\bm L^{(n)} \in \bbR{ (I_n-1) \times I_n}$ is a smoothness constraint matrix, typically defined as
\begin{align}
\bm L^{(n)} :=
\begin{pmatrix}
1 & -1 &        &        &    \\
  &  1 & -1     &        &    \\
  &    & \ddots & \ddots &    \\
  &    &        &  1     & -1 \\
\end{pmatrix}. \label{matrix:P}
\end{align}
The first term, $ || \ten{X} - \ten{Z} ||_F^2 $, of the objective function in \eqref{criterion:SmCPTC} represents the mean squared error (MSE) between the values of the observed entries $\ten{T}_\Omega$ and the PD model $\ten{Z}_\Omega$, because $\ten{X}_\Omega = \ten{T}_\Omega$ and $\ten{X}_{\bar{\Omega}} = \ten{Z}_{\bar{\Omega}}$.
Thus, the minimization of the first term of the objective function in \eqref{criterion:SmCPTC} provides a PD of a given tensor $\ten{T}$.

The second term of the objective function in \eqref{criterion:SmCPTC} is a penalty term, which assures smooth component vectors $\bm u_r^{(n)}$.
Note that we have $|| \bm L^{(n)} \bm u_r^{(n)} ||_p^p = \sum_{i=1}^{I_n - 1} | \bm u_r^{(n)}(i) - \bm u_r^{(n)}(i+1) |^p $, and the minimization of this non-smoothness measure enforces the smoothness of individual feature vectors.
When $p=1$, the constraint term becomes the total variation (TV), and when $p=2$ it becomes the quadratic variation (QV).
In contrast to LTVNN \cite{dai2009physics,han2014linear}, which imposes the smoothness constraint into the surface (output matrix), our approach
imposes the smoothness constraint into the cause (latent components).
The details of the difference between our approach and other existing smoothing methods are discussed in Section~\ref{sec:novelty}.

Note that the $g_r^2$ terms are scaling factors for the smoothness constraint terms in \eqref{criterion:SmCPTC}.
This allows us to adaptively enforce the different levels of smoothness into different components.
For image completion, the proposed method decomposes an image adaptively into a strong smooth background and a weaker smooth foreground.

\begin{algorithm}[t]
\caption{FR-SPC Algorithm} \label{algo:SmCPTC}
\begin{algorithmic}[1]
\STATE {\bf input:} $\ten{T}$, $\Omega$, $R$, $p$ and $\bm \rho$.
\STATE {\bf initialize:} $\{g_r, \{\bm u_r^{(n)} \in \mathbb{U}^{I_n} \}_{n=1}^N\}_{r=1}^R$, randomly;
\STATE Construct matrix $\bm L^{(n)}$ by \eqref{matrix:P}; $\forall n \in \{1,...,N\}$
\STATE $\ten{X}_\Omega \leftarrow \ten{T}_\Omega$;
\STATE $\ten{X}_{\bar{\Omega}} \leftarrow ( \sum_{r=1}^R g_r \bm u_r^{(1)} \circ \bm u_r^{(2)} \circ \dots \circ \bm u_r^{(N)} )_{\bar\Omega}$;
\STATE $\ten{E} = \ten{X} - \sum_{r=1}^R g_r \bm u_r^{(1)} \circ \bm u_r^{(2)} \circ \dots \circ \bm u_r^{(N)}$;
\REPEAT
  \FOR{$r = 1, ..., R$}
    \STATE $\ten{Y}_r \leftarrow \ten{E} + g_r \bm u_r^{(1)} \circ \bm u_r^{(2)} \circ \dots \circ \bm u_r^{(N)} $;
    \FOR{$n = 1, ..., N$}
      \STATE $\bm u_r^{(n)} \leftarrow \argmin_{ \bm u \in \mathbb{U}^{I_n} } F_p^{(r,n)}(\bm u) $;
    \ENDFOR
    \STATE $g_r \leftarrow  \frac{\langle \ten{Y}_r, \bm u_r^{(1)} \circ \bm u_r^{(2)} \circ \dots \circ \bm u_r^{(N)} \rangle}{(1 + \sum_{n=1}^N \rho^{(n)} || \bm L^{(n)} \bm u_r^{(n)} ||_p^p )}$;
    \STATE $\ten{E} \leftarrow \ten{Y}_r - g_r \bm u_r^{(1)} \circ \bm u_r^{(2)} \circ \dots \circ \bm u_r^{(N)} $;
    \STATE $\ten{E}_{\bar\Omega} \leftarrow 0$;
  \ENDFOR
\UNTIL{Change of $||\ten{E}||_F^2$ after the next iteration is sufficiently small.}
\STATE $\ten{Z} \leftarrow \sum_{r=1}^R g_r \bm u_r^{(1)} \circ \bm u_r^{(2)} \circ \dots \circ \bm u_r^{(N)}$;
\STATE $\ten{X}_{\bar\Omega} \leftarrow \ten{Z}_{\bar\Omega}$;
\STATE {\bf output:} $\ten{X}$, $\ten{Z}$.
\end{algorithmic}
\end{algorithm}

\subsection{Derivation of the FR-SPC Algorithm}
We solve the optimization problem \eqref{criterion:SmCPTC} using the hierarchical alternating least squares (HALS) approach \cite{cichocki2007hierarchical,cichocki2009nonnegative}.
The HALS algorithm, included in block coordinate descent scheme, considers `feature-wise' update which allows us to treat each unit-norm constraint separately.
According to the HALS approach, we consider the minimization of the following local cost functions:
\begin{align}
 \mathop{\text{minimize}}_{g_r,\bm u_r^{(1)},...,\bm u_r^{(N)}} &\ \ \frac{1}{2}|| \ten{Y}_r - \ten{Z}_r ||_F^2 + \frac{g_r^2}{2} \sum_{n=1}^N \rho^{(n)} || \bm L^{(n)} \bm u_r^{(n)} ||_p^p, \notag \\
 \text{s.t. }&\ \ \ten{Z}_r =  g_r \bm u_r^{(1)} \circ \bm u_r^{(2)} \circ \dots \circ \bm u_r^{(N)} \notag \\
             &\ \ [\ten{Y}_r]_\Omega = \ten{T}_\Omega^{(r)}, \ [ \ten{Y}_r ]_{\bar{\Omega}} = [ \ten{Z}_r ]_{\bar{\Omega}}, \label{criterion:SmCPTC:local} \\
             &\ \ ||\bm u_r^{(n)}||_2=1, \forall n \in \{1,...,N\},  \notag 
\end{align}
where $\ten{Y}_r = \ten{X} -  \sum_{i \neq r} g_i \bm u_i^{(1)} \circ \bm u_i^{(2)} \circ \dots \circ \bm u_i^{(N)} $ and $\ten{T}_\Omega^{(r)} = \ten{T}_\Omega - [ \sum_{i \neq r} g_i \bm u_i^{(1)} \circ \bm u_i^{(2)} \circ \dots \circ \bm u_i^{(N)} ]_\Omega $.
The local-problem \eqref{criterion:SmCPTC:local} only involves the $r$th components of PD with a fixed $\ten{X}$.

In order to solve \eqref{criterion:SmCPTC:local}, we update $\bm u_r^{(1)}, \bm u_r^{(2)}, ..., \bm u_r^{(N)}$, $g_r$, and reset $\ten{Y}_r$ sequentially as follows:
\begin{align}
  \bm u_r^{(n)} &\leftarrow \argmin_{\bm u \in \mathbb{U}^{I_n}} F_p^{(r,n)}(\bm u),\label{criterion:SmCPTC:local:u} \\
  g_r &\leftarrow \argmin_{g} F_p^{(r)}(g), \label{criterion:SmCPTC:local:g}\\
  [\ten{Y}_r]_{\bar{\Omega}} &\leftarrow  [g_r \bm u_r^{(1)} \circ \bm u_r^{(2)} \circ \dots \circ \bm u_r^{(N)}]_{\bar{\Omega}},
\end{align}
where $F_p^{(r)}(g_r) = F_p^{(r,n)}(\bm u_r^{(n)}) = \frac{1}{2}|| \ten{Y}_r - g_r  \bm u_r^{(1)} \circ \bm u_r^{(2)} \circ \dots \circ \bm u_r^{(N)} ||_F^2 +  \frac{g_r^2}{2} \sum_{n=1}^N \rho^{(n)} || \bm L^{(n)} \bm u_r^{(n)} ||_p^p $ for all $n$ and $r$, and $\mathbb{U}^{I_n} := \{ \bm u \in \bbR{I_n}\ |\  ||\bm u||_2 = 1 \}$ is a sub-set of unit vectors.

\subsubsection{Update Rule for $\bm u_r^{(n)}$}
The problem \eqref{criterion:SmCPTC:local:u} can be characterized as a unit-norm constrained optimization.
When $p=2$, objective function is a quadratic function with respect to $\bm u$ and it can be formulated as spherical constrained quadratic optimization \cite{sorensen1997minimization}.
When $p=1$, the objective function is non-differentiable.
Fortunately, many methods for solving this problem have been studied, such as sub-gradient methods \cite{bertsekas2015convex}.
To treat both cases (i.e., $p=1$ and $2$) equally, in this paper, we consider the gradient (or sub-gradient) based coefficient normalization update method \cite{owsley1978adaptive,douglas2000gradient}.
Although it does not guarantee global convergence generally, sticking into local minima of sub-optimization problem is not critical issue because the objective of sub-optimization is to decrease global cost function rather than to obtain strict solution of itself.
As other options, several optimization schemes can be also applied such as Lagrange multiplier method \cite{courant1966methods}, tangent gradient \cite{krasulina1970method} and optimization method on manifolds \cite{absil2009optimization}.
In order the simplify the notation, we denote by $\bm u_k$ and $\bm v_k$ the $k$-th updates of $\bm u_r^{(n)}$ and $\partial F_p^{(r,n)}(\bm u_r^{(n)})$ in the iterative algorithm, respectively, where $\partial F_p^{(r,n)}(\cdot)$ is a gradient (or sub-gradient) of the objective function.
Thus, the update $\bm u_{k+1}$ is given by
\begin{align}
  \bm u_{k+1} = \frac{\bm u_k - \alpha \bm v_k}{ \sqrt{1 - 2\alpha \bm u_k^T \bm v_k + \alpha^2 \bm v_k^T \bm v_k} }, \label{gradient_unit_norm_update}
\end{align}
where $\alpha$ a step size parameter, which should be tuned so that $ F_p^{(r,n)}(\bm u_{k+1}) \leq F_p^{(r,n)}(\bm u_k) $.
This update rule can be considered as enforcing a sub-gradient descent on the hyper-spherical surface $||\bm u_k||_2=||\bm u_{k+1}||_2=1$.
We iterate \eqref{gradient_unit_norm_update} until convergence is achieved.

Next, we will demonstrate the gradient (or sub-gradient) of the objective function.
The objective function $F_p^{(r,n)}(\bm u_k)$ can be simplified as
\begin{align}
  \frac{g_r^2}{2} \rho^{(n)} || \bm L^{(n)} \bm u_k ||_p^p - g_r \bm u_k^{T} \bm y_r^{(n)} + \frac{1}{2}g_r^2  \bm u_k^{T} \bm u_k,
\end{align}
where $\bm y_r^{(n)} := \text{vec}(\ten{Y}_r \times_1 \bm u_r^{(1)T} \times_2 \cdots \times_{n-1} \bm u_r^{(n-1)T} \times_{n+1} \bm u_r^{(n+1)T} \times_{n+2} ... \times_N \bm u_r^{(N)T})$.
Thus, the gradient (or sub-gradient) of the objective function is given by
\begin{align}
&\partial F_p^{(r,n)}(\bm u_k) = \notag \\
&\left\{
 \begin{array}{ll}
  \frac{g_r^2}{2} \rho^{(n)} \bm L^{(n)T} \mathrm{SGN}(\bm L^{(n)} \bm u_k) - g_r \bm y_r^{(n)} + g_r^2  \bm u_k & (p=1) \\
  g_r^2 \rho^{(n)} \bm L^{(n)T} \bm L^{(n)} \bm u_k - g_r \bm y_r^{(n)} + g_r^2  \bm u_k & (p=2)
 \end{array} \right. ,
\end{align}
where the vector function $\mathrm{SGN}$ is defined by
\begin{align}
 &\mathrm{SGN}(\bm x) = [\text{sgn}(x_1), \text{sgn}(x_2), ..., \text{sgn}(x_J)]^T, \\
 &\text{sgn}(x_j) = \left\{ \begin{array}{ll} 1 & (x_j > 0) \\ 0 & (x_j=0) \\ -1 & (x_j<0) \end{array} \right. ,
\end{align}
for any $\bm x \in \bbR{J}$.
This definition of $\mathrm{SGN}(\bm x)$ provides the most unbiased sub-gradient of $||\bm x||_1$.

\subsubsection{Update Rule for $g_r$}
Because the problem in \eqref{criterion:SmCPTC:local:g} is an example of unconstrained quadratic optimization, the unique solution can be obtained analytically.
The objective function can be simplified as
\begin{align}
  \frac{1}{2}g_r^2 & - g_r \langle \ten{Y}_r, \bm u_r^{(1)} \circ \bm u_r^{(2)} \circ \dots \circ \bm u_r^{(N)} \rangle \notag \\
  & + \frac{1}{2}g_r^2 \sum_{n=1}^N \rho^{(n)} || \bm L^{(n)} \bm u_r^{(n)} ||_p^p.
\end{align}
Thus, the update rule is given by
\begin{align}
  g_r \leftarrow  \frac{\langle \ten{Y}_r, \bm u_r^{(1)} \circ \bm u_r^{(2)} \circ \dots \circ \bm u_r^{(N)} \rangle}{(1 + \sum_{n=1}^N \rho^{(n)} || \bm L^{(n)} \bm u_r^{(n)} ||_p^p )}.
\end{align}
Finally, the FR-SPC optimization scheme can be summarized as shown in Algorithm \ref{algo:SmCPTC}.
According to \cite{yu2014parallel}, several iterations of inner-loop of 10-13th lines in Algorithm~\ref{algo:SmCPTC} may accelerate convergence speed for $N = 2$, however it is not necessary for $N \geq 3$ \cite{karlsson2014parallel}.

\begin{algorithm}[t]
\caption{Algorithm for Estimation of Optimal Number of Components $R$} \label{algo:R_simple}
\begin{algorithmic}[1]
\STATE {\bf input:} $\ten{T}$, $\Omega$, $p$, $\bm \rho$, and SDR.
\STATE $\varepsilon \leftarrow 10^{(-\text{SDR}/10)} ||\ten{T}_{\Omega}||_F^2$;
\STATE $R \leftarrow 1$;
\REPEAT 
  \STATE $[\ten{Z}, \ten{X}] \leftarrow \text{FR-SPC}(\ten{T}, \Omega, R, p, \bm \rho)$;
  \STATE $R \leftarrow R + 1$;
\UNTIL{ $|| \ten{Z}_\Omega - \ten{T}_\Omega ||_F^2 \leq \varepsilon $ }
\STATE {\bf output:} $\ten{Z}$, $\ten{X}$.
\end{algorithmic}
\end{algorithm}

\begin{algorithm}[t]
\caption{SPC  Algorithm (accelerated version of Algorithm \ref{algo:R_simple})} \label{algo:EXTRA}
\begin{algorithmic}[1]
\STATE {\bf input:} $\ten{T}$, $\Omega$, $p$, $\bm \rho$, SDR, and $\nu$.
\STATE $\varepsilon \leftarrow 10^{(-\text{SDR}/10)} ||\ten{T}_{\Omega}||_F^2$;
\STATE $\ten{X}_\Omega \leftarrow \ten{T}_\Omega$;\  $\ten{X}_{\bar\Omega} \leftarrow \text{average of } \ten{T}_\Omega $;
\STATE Construct matrix $\bm L^{(n)}$ by \eqref{matrix:P}; $\forall n \in \{1,...,N\}$
\STATE $R \leftarrow 1$;
\STATE Initialize $\{\bm u_R^{(n)} \in \mathbb{U}^{I_n} \}_{n=1}^N$, randomly;
\STATE $g_R \leftarrow \langle \ten{X}, \bm u_R^{(1)} \circ \bm u_R^{(2)} \circ \dots \circ \bm u_R^{(N)} \rangle$;
\STATE $\ten{E} = \ten{X} - \sum_{r=1}^R g_r \bm u_r^{(1)} \circ \bm u_r^{(2)} \circ \dots \circ \bm u_r^{(N)}$;
\STATE $\ten{E}_{\bar\Omega} = 0$;
\STATE $t \leftarrow 0$;
\STATE $\mu_t \leftarrow ||\ten{E}||_F^2$;
\REPEAT
  \STATE Update $\{\{\bm u_r^{(n)}\}_{n=1}^N, g_r \}_{r=1}^R$ by the FR-SPC algorithm (8th-16th lines);
  \STATE $\mu_{t+1} \leftarrow ||\ten{E}||_F^2$;
  \IF{ $\frac{|\mu_t - \mu_{t+1}|}{|\mu_{t+1} - \varepsilon|} \leq \nu$ }
    \STATE $R \leftarrow R+1$;
    \STATE Initialize $\{\bm u_R^{(n)} \in \mathbb{U}^{I_n} \}_{n=1}^N$, randomly;
    \STATE $g_R \leftarrow \langle \ten{E}, \bm u_R^{(1)} \circ \bm u_R^{(2)} \circ \dots \circ \bm u_R^{(N)} \rangle$;
    \STATE $\ten{E} \leftarrow \ten{E} - g_R \bm u_R^{(1)} \circ \bm u_R^{(2)} \circ \dots \circ \bm u_R^{(N)}$;
    \STATE $\ten{E}_{\bar\Omega} \leftarrow 0$;
  \ENDIF
  \STATE $t \leftarrow t+1$;
\UNTIL{$\mu_t \leq \varepsilon$}
\STATE $\ten{Z} \leftarrow \sum_{r=1}^R g_r \bm u_r^{(1)} \circ \bm u_r^{(2)} \circ \dots \circ \bm u_r^{(N)}$;
\STATE $\ten{X}_{\bar\Omega} \leftarrow \ten{Z}_{\bar\Omega}$;
\STATE {\bf output:} $\ten{X}$, $\ten{Z}$.
\end{algorithmic}
\end{algorithm}

\begin{figure}[t]
\centering
\includegraphics[width = \textwidthh]{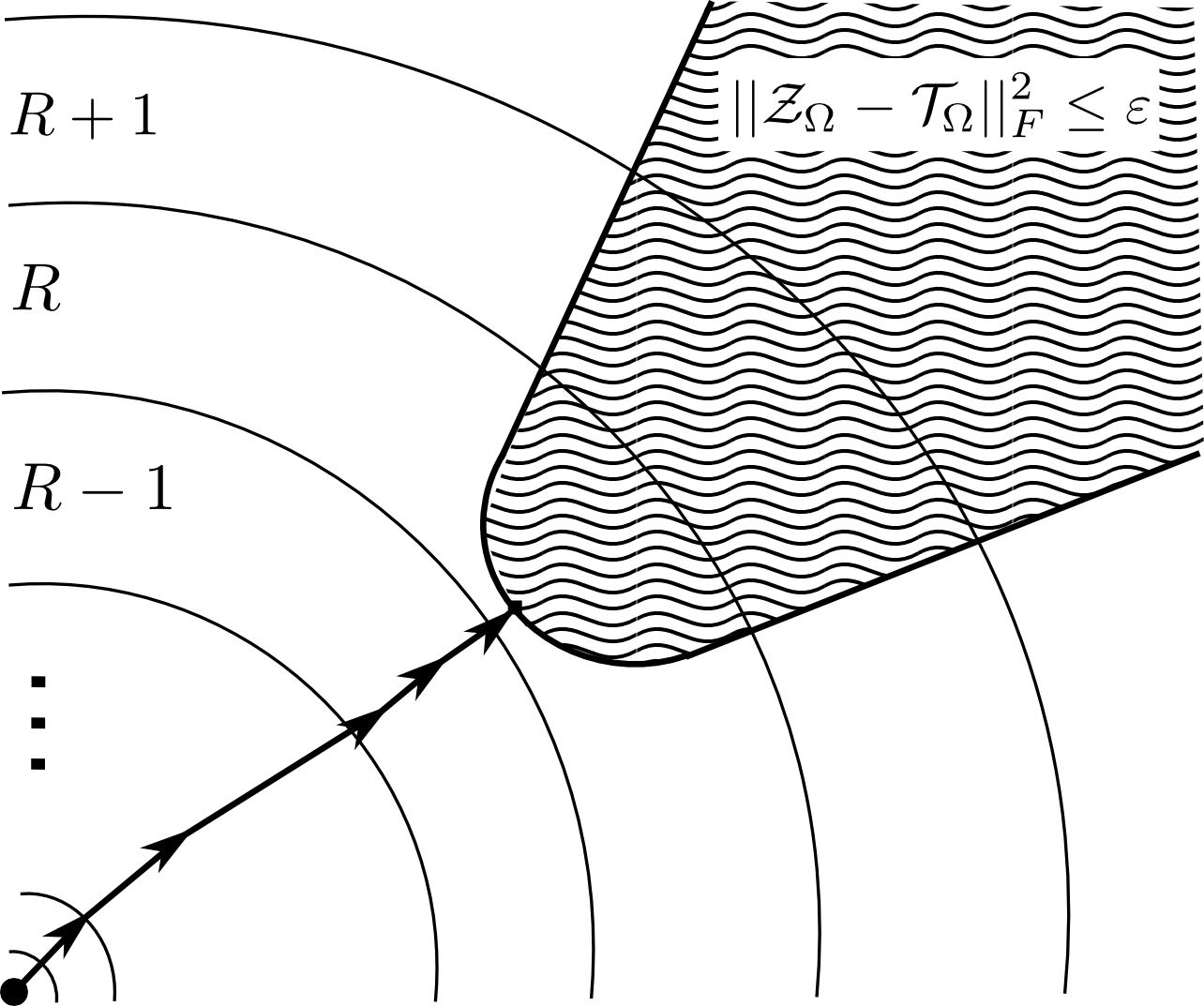}
\caption{Concept of optimization process in the SPC algorithm. An area $R$ between arcs indicates the space of the tensor that is spanned by the components of $R$ in the smooth PD model. An area that is filled with waves indicates the space of the tensor satisfying the condition $|| \ten{Z}_\Omega - \ten{T}_\Omega ||_F^2 \leq \varepsilon$. Concatenated arrows show the optimization process of the SPC algorithm, which gradually increases the number of components $R$ from a rank-1 tensor to its end point, when the criterion is satisfied.}\label{fig:optimization}
\end{figure}

\subsection{Model Selection for the Number of Components $R$}\label{sec:rank_det}
The key problem of FR-SPC is choosing the optimal number of components $R$.
The PD model with too small an $R$ is not able to fit the data, and the PD model with too large an $R$ may result in over-fitting problems.

In order to estimate an optimal value of $R$, we gradually increase $R$ until we achieved the desired fit by formulating the following optimization problem:
\begin{align}
\mathop{\text{minimize}}_R \ \ &R, \label{optimize:R}\\
\text{s.t. } \ \ & || \ten{Z}_\Omega - \ten{T}_\Omega ||_F^2 \leq \varepsilon, \notag\\
                 & \ten{Z} \in \mathbb S (R,p,\bm \rho), \notag
\end{align}
where $\mathbb S(R,p,\bm \rho)$ is defined as a set of all possible solution tensors via FR-SPC problem \eqref{criterion:SmCPTC:local} with $R$, $p$, and $\bm \rho$ for any input tensors, in other words $\mathbb S(R,p,\bm \rho)$ is a tensor space spanned by $R$ smooth components, and $\varepsilon$ is an error bound parameter.
The criterion \eqref{optimize:R} allows us to finds a tensor $\ten{Z}$ based on the smooth PD model, with the minimum number of smooth components that guarantees a sufficient accuracy to fit the input tensor $\ten{T}$.
In order to guarantee that the signal to distortion ratio (SDR) is bounded within a specific threshold, we can define the error bound as $\varepsilon = 10^{(-\text{SDR}/10)} ||\ten{T}_{\Omega}||_F^2$.
Note that $\mathop{\min}_{\ten{Z} \in \mathbb S(R,p,\bm \rho)}\{ || \ten{Z}_\Omega - \ten{T}_\Omega ||_F^2 \}$ is a monotonically non-increasing function with respect to $R$.
Algorithm~\ref{algo:R_simple} is an implementation to solve \eqref{optimize:R}, and it allows us to obtain a good estimation for $R$.
However, this algorithm is quite time consuming.
If the current value of $R$ is too small to fit the PD model sufficiently well to the given data tensor during the iteration process, then we can stop the algorithm for the current $R$, increase as $R \leftarrow R + 1$, and run the algorithm again for the new increased $R$.
In this procedure, we propose switching to an increased $R$ at an early stage of the iterations if the following condition is met:
\begin{align}
 \frac{ | \mu_{t} - \mu_{t+1} |}{| \mu_{t+1} - \varepsilon |} < \nu, \label{skipping_condition}
\end{align}
where $\mu_{t} = || \ten{Z}^{t}_\Omega - \ten{T}_\Omega ||_F^2$, $\ten{Z}^{t}$ denotes the PD model at the iteration step $t$, and $\nu > 0$ is a stopping threshold (typically, $\nu=0.01$).
The left hand side of \eqref{skipping_condition} is a measure of the convergence speed, so this condition means that when the convergence speed becomes substantially slow, we stop the iteration procedure for the current $R$.
By incorporating this simple stopping criterion, we finally arrive at an improved and considerably accelerated algorithm for the automatic determination of the number of components $R$ of FR-SPC.
We call this method simply ``smooth PARAFAC tensor completion with TV and QV minimization'' (SPC-TV and SPC-QV), and it is summarized in Algorithm~\ref{algo:EXTRA}.
Fig.~\ref{fig:optimization} presents the concept of the optimization process of the SPC algorithm, and its details are provided in Section~\ref{sec:disc}.
Our algorithms have been implemented in MATLAB, and are available for readers at \url{https://sites.google.com/site/yokotatsuya/home/software}.

Because the best low-rank approximation problem for smooth PD model may be often ill-posed in the same way as for unconstrained PD model \cite{de2008tensor}, our algorithm allows some redundant terms in smooth PD model.
Focusing on ``good approximation'' rather than ``minimum rank decomposition'', although estimation of number of components for smooth PD model is difficult, its accuracy does not need to be so high if the reconstructed tensor is well approximated in practice.

\begin{figure}[t]
\centering
\includegraphics[width = \textwidthh]{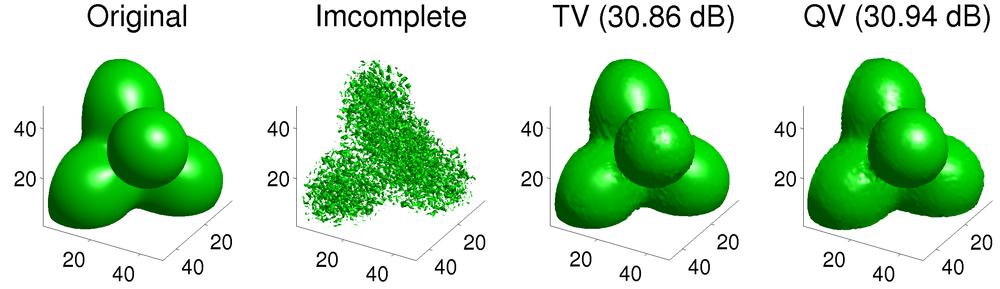}
\caption{Iso-surface visualization of synthetic data and completion results using SPC-TV and SPC-QV.}\label{fig:toy}
\end{figure}

\begin{figure}[t]
\centering
\subfigure[TV]{\includegraphics[width = \textwidthhh]{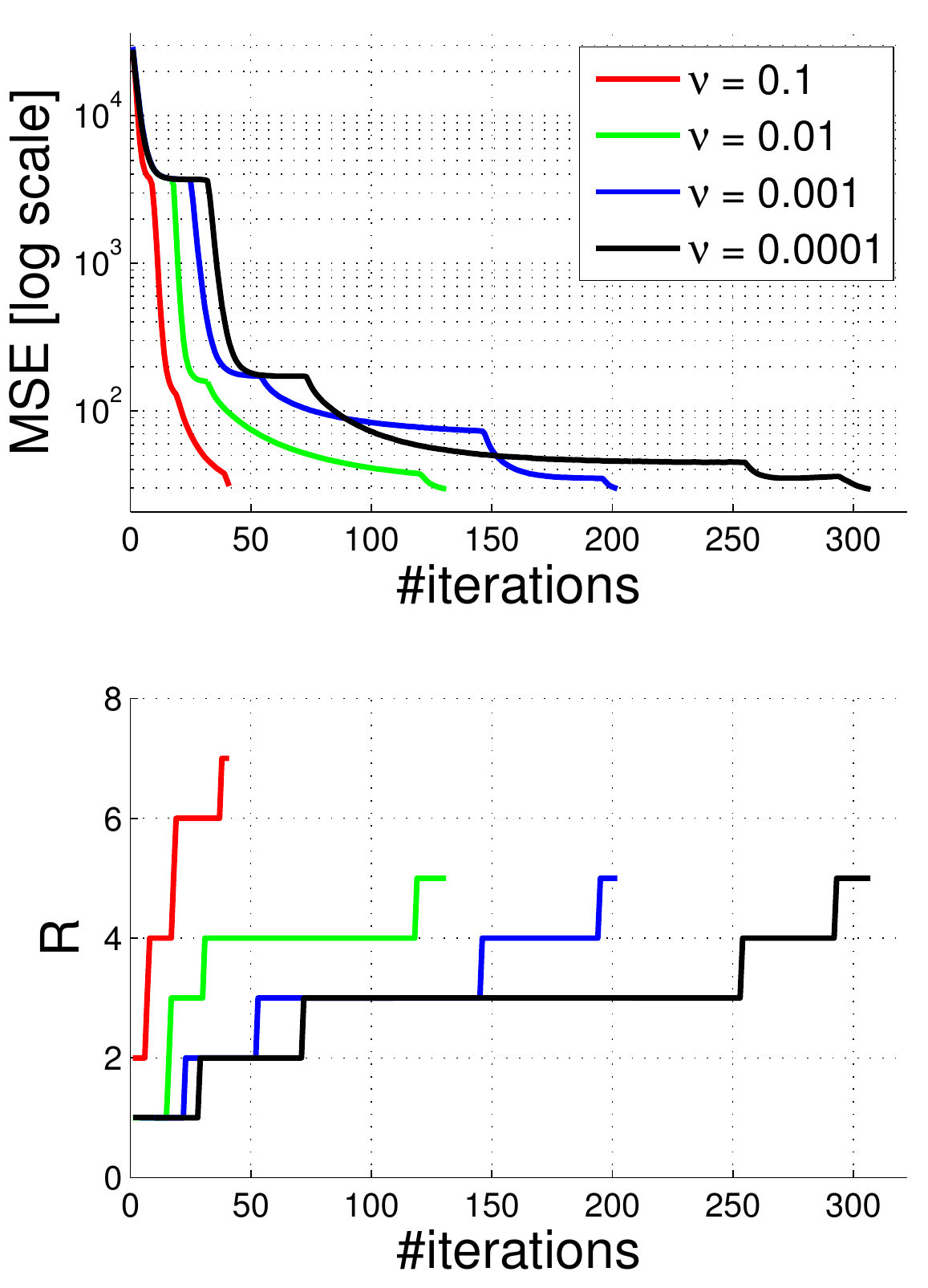}}
\subfigure[QV]{\includegraphics[width = \textwidthhh]{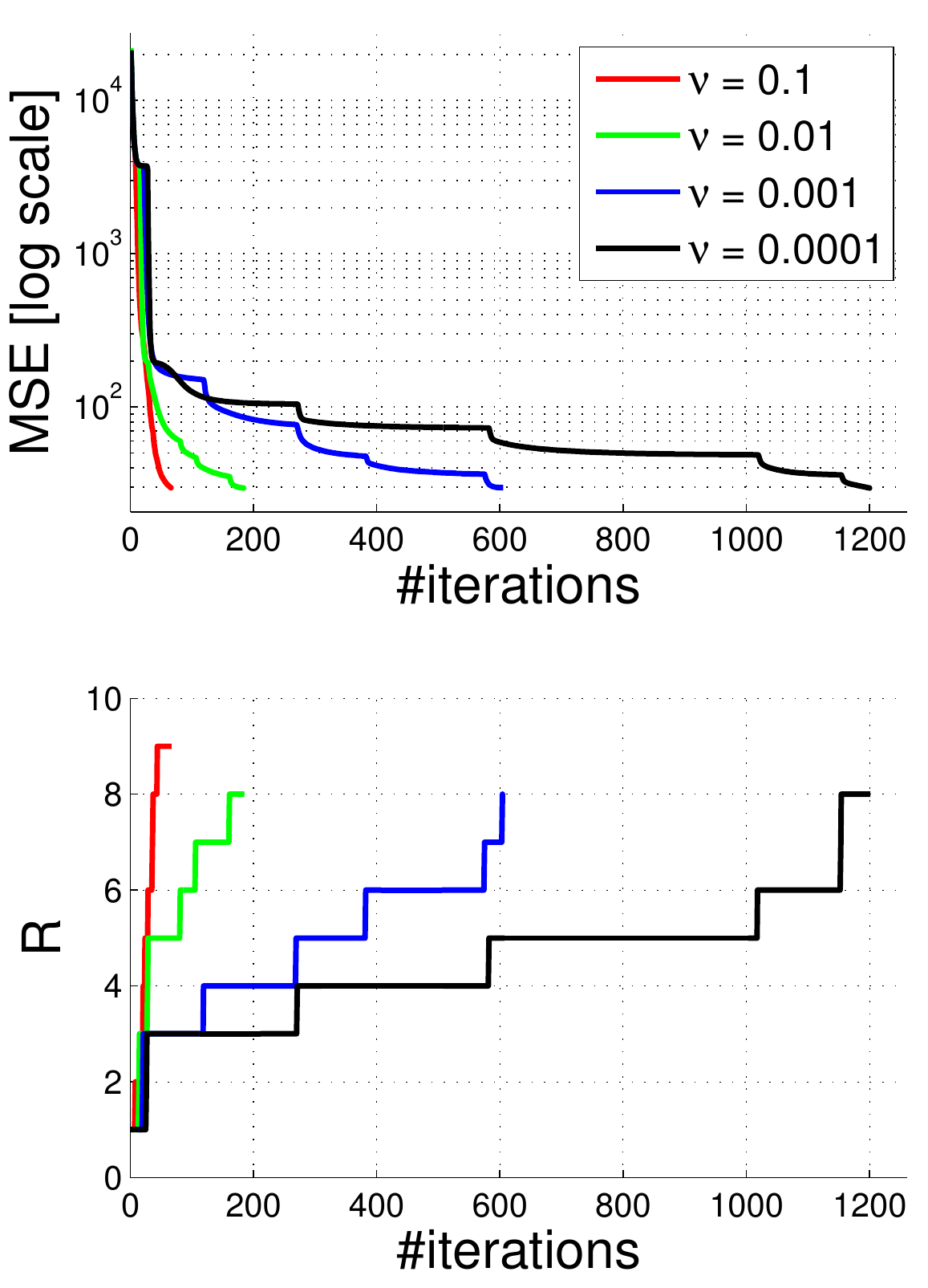}}
\caption{Convergence of mean squared error between $\ten{Z}_\Omega$ and $\ten{T_\Omega}$, and number of components $R$ as a function of the number of iterations in the SPC algorithm using the synthetic data with missing voxels rate of 80\%. We performed simulations for various values of the stopping threshold $\nu \in \{0.1, 0.01, 0.001, 0.0001 \}$. Other parameters were set as SDR$=30$ dB, $\bm\rho = [0.01, 0.01, 0.01]$ for TV smoothing, and $\bm\rho=[1.0, 1.0, 1.0]$ for QV smoothing.}\label{fig:convergence}
\end{figure}

\section{Experimental Results}\label{sec:experiments}
In the experiments to be described here, we apply our proposed algorithms to one synthetic and several real-world visual datasets, to demonstrate the sensitivity of the parameters $\nu$ and $\bm \rho$, and the advantages of our proposed algorithms compared with the existing state-of-the-art methods LTVNN \cite{han2014linear}, HaLRTC \cite{liu2013tensor}, STDC \cite{chen2014simultaneous}, and FBCP-MP \cite{zhao2015bayesian}.

\subsection{Convergence Properties Using a Synthetic Third-Order Tensor }
Fig.~\ref{fig:toy} shows a visualization of the iso-surface of a synthetic 3rd-order tensor, and its completion results by achieved by applying the SPC algorithm with TV and QV constraints.
The synthetic tensor was constructed using a combination of four multi-dimensional Gaussian functions, and $80\%$ of the voxels were randomly removed in its incomplete tensor.

Fig.~\ref{fig:convergence} shows the plots of the mean of squared errors (MSE) between the tensors $\ten{Z}_\Omega$ and $\ten{T}_\Omega$, and the number of estimated components R with respect to number of iterations of the SPC algorithm, with $\bm\rho = [0.01, 0.01, 0.01]$ for TV smoothing and $\bm\rho=[1.0, 1.0, 1.0]$ for QV smoothing. In addition, we set SDR$=30$ dB and applied various switching threshold values of $\nu \in \{0.1, 0.01, 0.001, 0.0001\}$.
We used the synthetic smooth third-order tensor with a random $80\%$ of voxels missing.
From Fig.~\ref{fig:convergence}, we can see the monotonic convergence of MSE and the change in the number of components of $R$ are closely related to the change of the MSE.
Note that the number of components, $R$, is updated when the convergence speed of the MSE becomes slow.
Furthermore, the algorithm is seen to require a much larger number of iterations for smaller values of $\nu$.
However, the results for the final number of components $R$ were the same for different cases with $\nu \geq 0.01$.
This means that for too small a value of $\nu$ the convergence of the algorithm is relatively slow.
According to our extensive experiments, $\nu=0.01$ is an acceptable default value.

\begin{figure}[t]
\centering
\includegraphics[width = 0.49\textwidth]{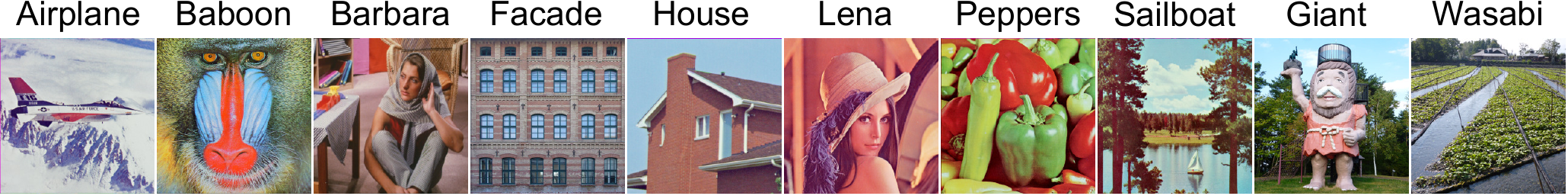}
\caption{Test images are $256 \times 256$ pixels color images, which consist of three layers of red, green, and blue color images.}\label{fig:test_images}
\vspace*{5mm}
\centering
\subfigure[Convergence curves]{\includegraphics[width=0.29\textwidth]{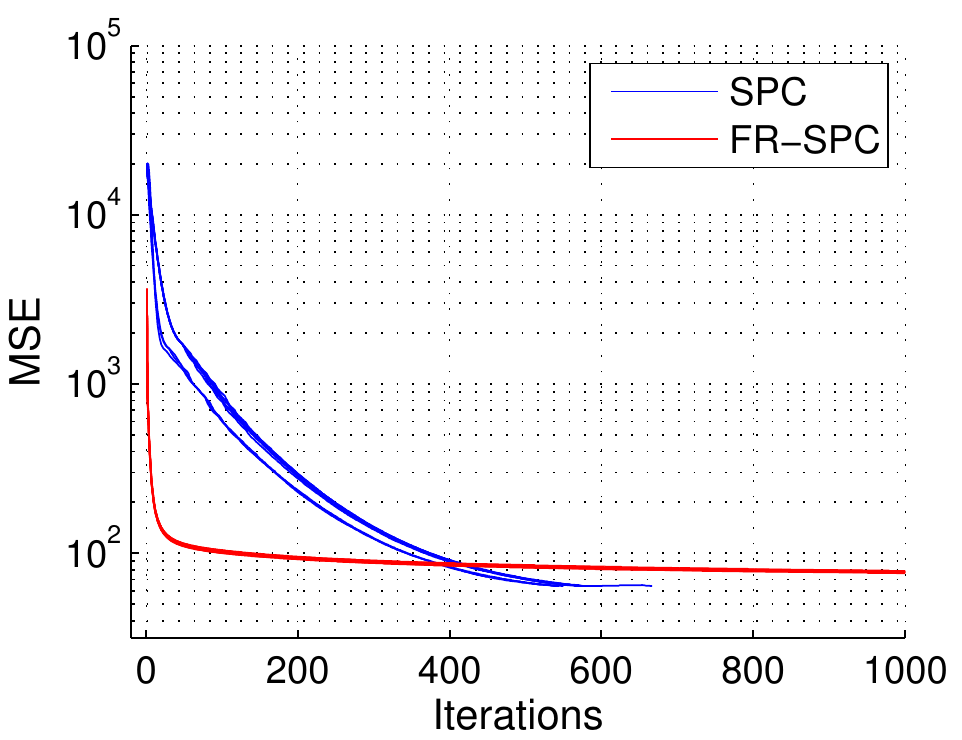}}
\subfigure[PSNR and SSIM statistics]{\includegraphics[width=0.19\textwidth]{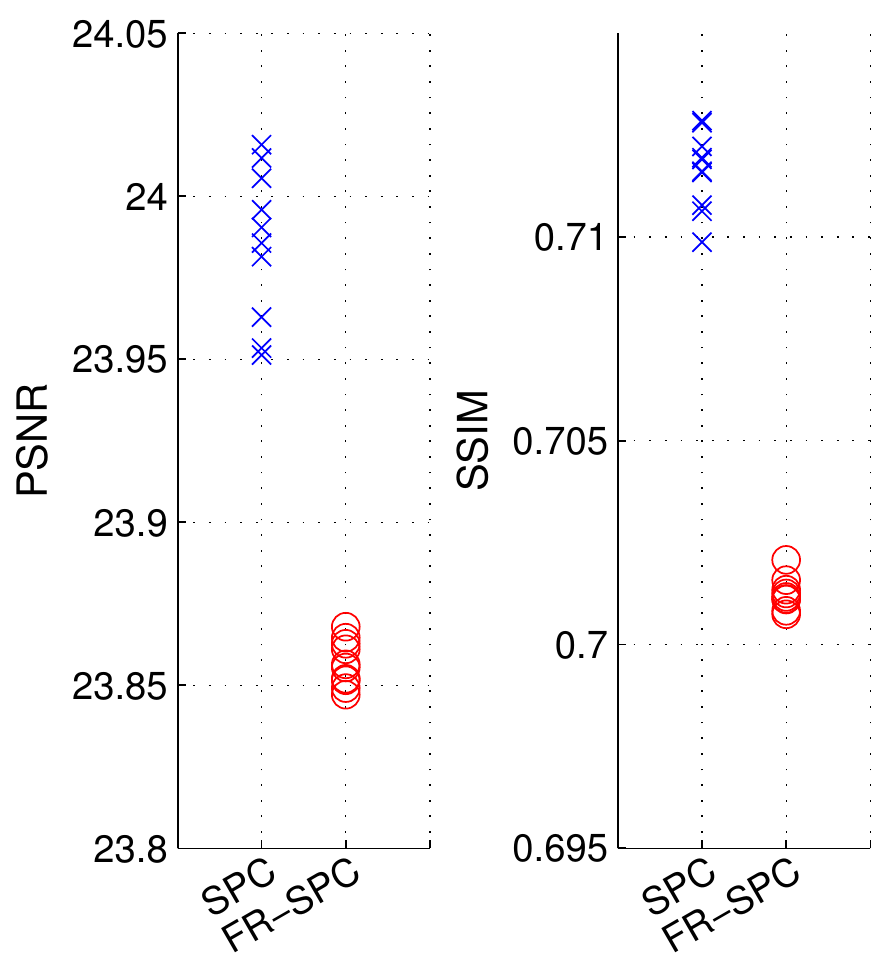}}
\caption{Comparison of the SPC and FR-SPC algorithms.}\label{fig:comp_fr}
\vspace*{5mm}
\centering
\includegraphics[width = 0.49\textwidth]{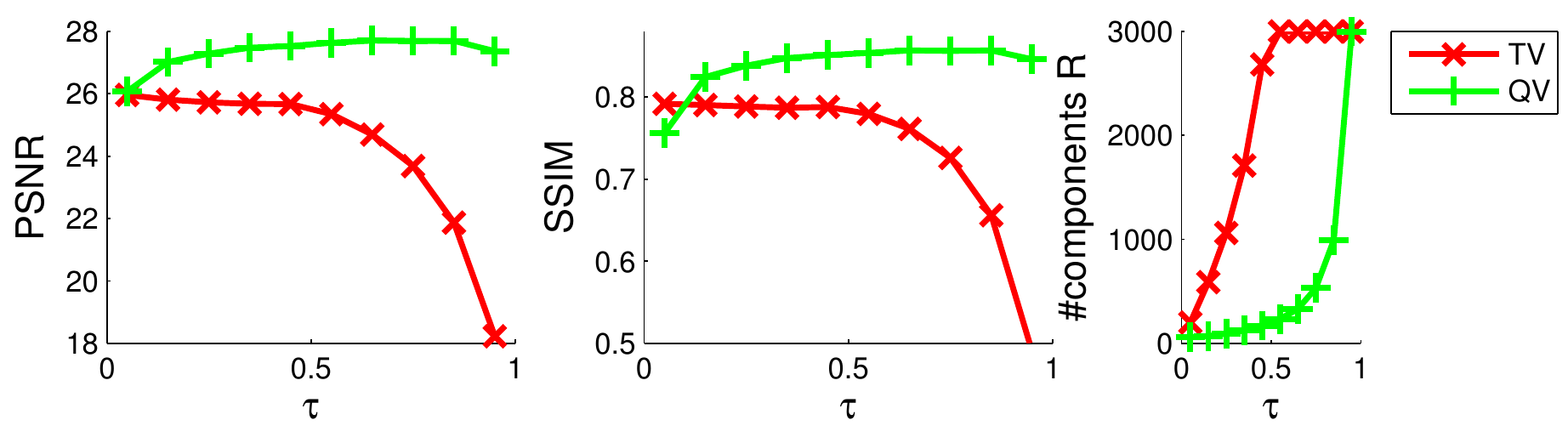}
\caption{Results of PSNR, SSIM, and the number of components of the SPC  algorithms, using the image ``Lena'' with a missing ratio of $80\%$. We tested various values of $\bm\rho = [\rho, \rho, 0]^T$, where $\rho$ is calculated by $\rho = \tau/(1-\tau)$. The algorithms were relatively insensitive to the wide variation of the parameters.}\label{fig:var_rho}
\end{figure}

\begin{figure*}[t]
\centering
\subfigure[Relative PSNR subtracted by LTVNN]{\includegraphics[width = 0.99\textwidth]{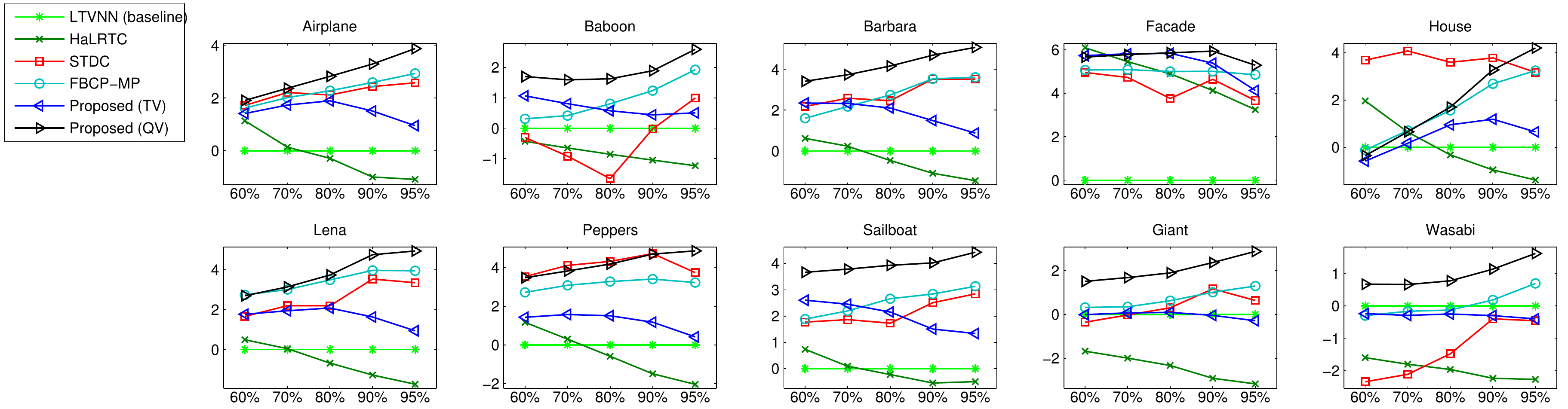}}
\subfigure[Relative SSIM subtracted by LTVNN]{\includegraphics[width = 0.99\textwidth]{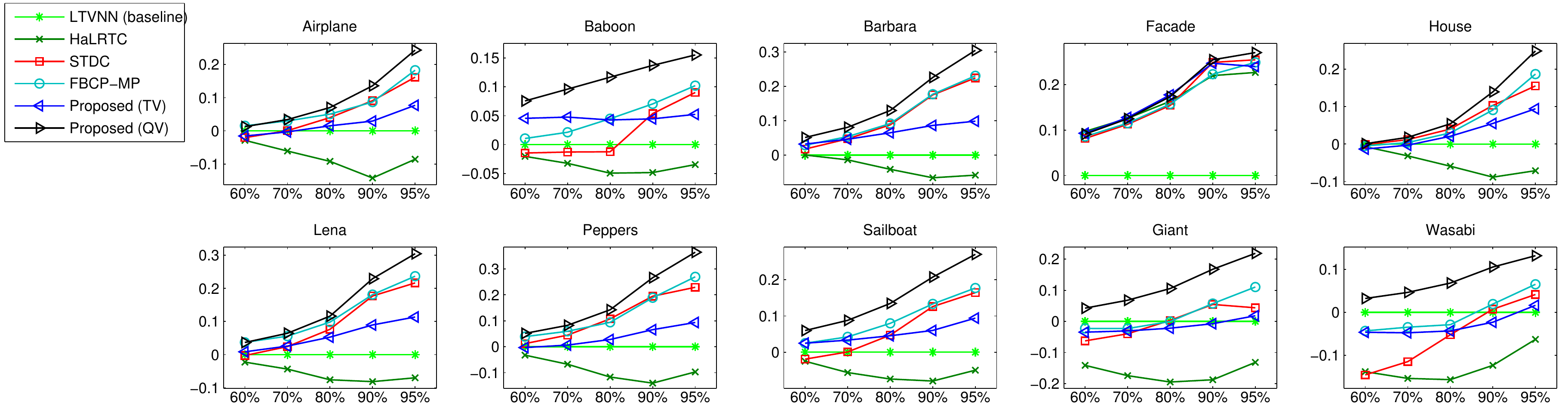}}
\caption{Comparison of performances (PSNR and SSIM subtracted by baseline (LTVNN)) for all benchmark images and for various missing rates (from 60\% to 95\%) obtained using the proposed methods (SPC-TV and SPC-QV) and the state-of-the-art algorithms LTVNN, HaLRTC, STDC, and FBCP-MP.}\label{fig:PSNR_SSIM}
\end{figure*}

\begin{figure*}[t]
\centering
\includegraphics[width = 0.95\textwidth]{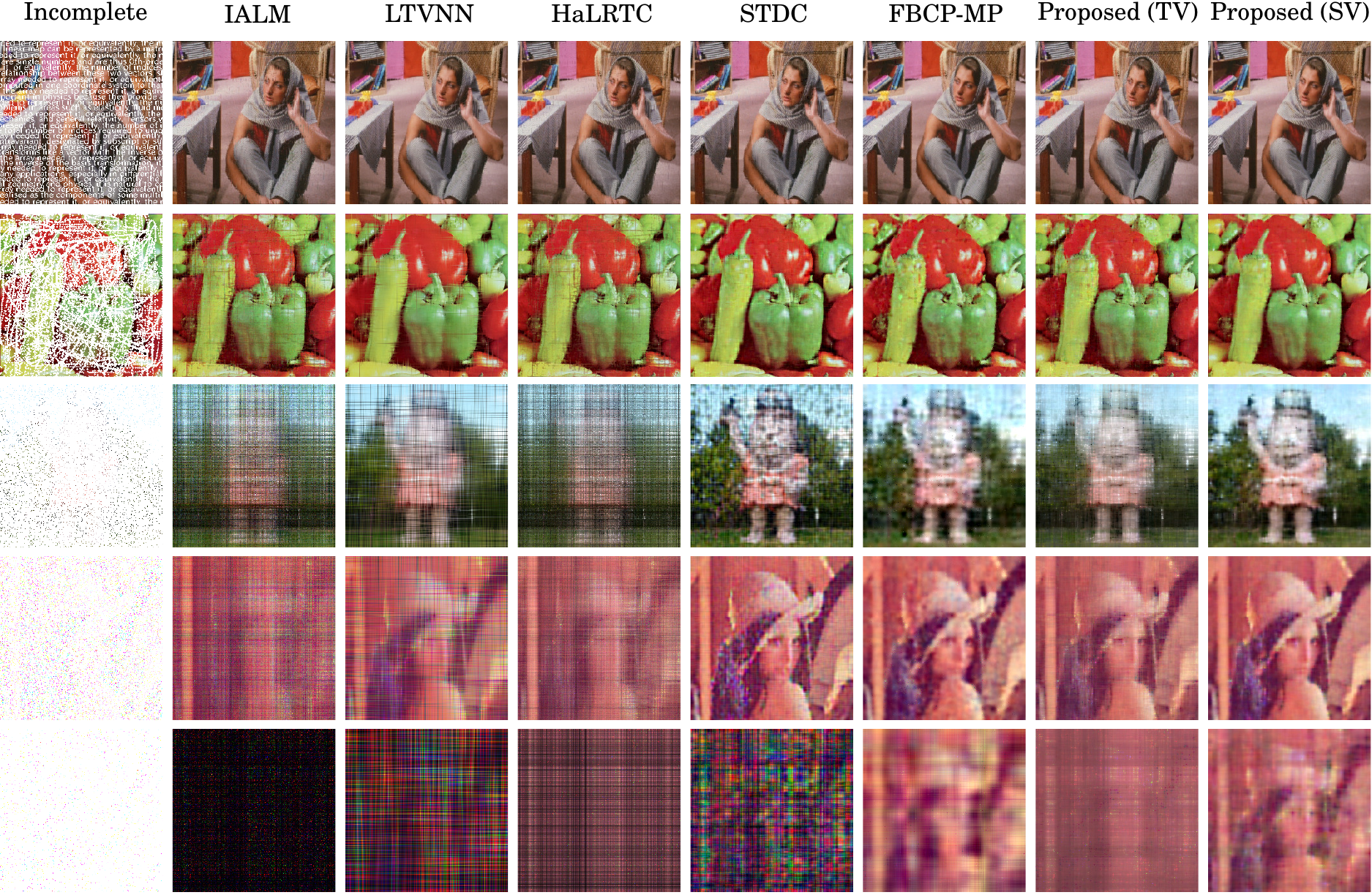}
\caption{Simulation results for various incomplete images (top to bottom): text masked image, scratched image, random dead pixels missing (95\%), and random voxels missing (95\% and 99\%). Incomplete images are shown in the first column, and reconstructed images are shown in the other columns.}\label{fig:selected_images}
\end{figure*}

\subsection{Color Image Completion}
In this experiment, we applied our proposed and existing methods to 10 color-images (Fig.~\ref{fig:test_images}).
The size of all color images is $256 \times 256 \times 3$, and we generated incomplete data by deleting elements of the images randomly, with several different missing ratios $\in \{60, 70, 80, 90, 95 \% \}$.
For the images ``Giant'' and ``Wasabi,'' all color elements of individual pixels (so called dead pixels) were deleted for all missing ratios.
In order to evaluate the quality of completion, we used the peak signal-to-noise ratio (PSNR) and structural similarity index (SSIM) \cite{wang2004image}.
PSNR is defined as $10\log_{10}(255^2/\text{MSE})$, and SSIM is defined by using the average and variance parameters of the local small blocks in an image.
As an improved evaluation measure, SSIM is usually considered as an improvement on PSNR, because SSIM evaluates both the error and a kind of ``local smoothness'' of images for visual quality.
In this experiment, we generally set $\text{SDR}=25$ dB and $\nu=0.01$.

\subsubsection{Convergence Behavior of SPC and FR-SPC Algorithms}
Fig.~\ref{fig:comp_fr} shows the convergence curves for PSNR and SSIM for 10 runs with random initializations in the SPC and FR-SPC algorithms, using ``Lena'' with a random $95\%$ of pixels missing, with $p=2$ and $\bm\rho = [0.5, 0.5, 0]$.
In both methods, the number of components is the same.
Interestingly, FR-SPC converged to local minima that are larger than the convergence points for SPC. Furthermore, the PSNR and SSIM values achieved by SPC were significantly better than those for the FR-SPC algorithm.
This implies that the ``rank-increasing'' approach results in an algorithm that converges to better local minima.

\subsubsection{Performances for Various Values of $\rho$}
Fig.~\ref{fig:var_rho} shows the PSNR, SSIM, and number of components $R$ for different values of the smoothing parameter $\bm\rho$.
A benchmark image, ``Lena,'' with a random $80\%$ of pixels missing, was used in this experiment.
We applied the SPC-TV and SPC-QV algorithms, with various values of $\bm\rho = [\rho, \rho, 0]$.
We defined $\rho := \tau/(1-\tau)$, and tested for a wide range of $\tau \in [0.05, 0.95]$.
In this experiment, we set the maximum values for $R$ as $3000$, because of the limitations of the memory.
We found that larger values of $\rho$ resulted in considerable increases in PSNR and SSIM in QV/TV smoothing. However, too large a $\rho$ substantially decreases the performances in TV smoothing.
Furthermore, a larger $\rho$ requires a larger number of components $R$ to achieve a more precise fit, and also a larger computational cost.
The SPC algorithm with TV smoothing required a larger number of components $R$ than QV smoothing, and usually the performances, in terms of PSNR and SSIM, were inferior in comparison with QV smoothing.
According to our results, QV smoothing performs, in general, better than TV smoothing for tensor completion.
We discuss our interpretation of these results in Section~\ref{sec:dics:TVQV}.

\subsubsection{Comparison with the State-of-the-Art Methods}
Fig.~\ref{fig:PSNR_SSIM} shows the results of tensor completions using all 10 benchmark images with various random pixel missing ratios using the proposed method with TV and QV smoothing, and compares these with the performances of four state-of-the-art methods.
The parameters for the SPC algorithms were set as $\bm\rho = [0.05, 0.05, 0]$ for TV smoothing, and $\bm\rho=[1.0, 1.0, 0]$ for QV smoothing.
The hyper-parameters for the other methods were optimally tuned manually.
Interestingly, LTVNN outperformed HaLRTC for several images with respect to SSIM. However, it is difficult to say whether LTVNN outperformed HaLRTC with respect to PSNR.
This implies that the smoothness constraint improves the structural similarity of images.
Thus, we can conclude that the smoothed methods (i.e., LTVNN, STDC, FBCP-MP, and the proposed method) are evaluated better by SSIM than PSNR.
The proposed method with TV smoothing was inferior to STDC and FBCP-MP in most cases (e.g., ``Peppers'' and ``House''). However, the proposed method with QV smoothing considerably outperformed the all of the existing methods for all of our benchmarks.

Fig.~\ref{fig:selected_images} presents several results for the completed images, using the existing methods and the proposed method.
We prepared a text masked image of ``Barbara,'' a scratched image of ``Peppers,'' an incomplete image with random pixels missing (dead pixels) of ``Giant,'' and two incomplete images with random voxels missing of ``Lena.''
In addition, for the first time we tested an extreme case where the ``$99\%$'' of pixels are missing, and such a large missing ratio may constitute an interesting challenge in this research field.
There were rather small difference between the performances of most of the methods in the first and second rows. However, we observed significant differences for the images  missing random pixels/voxels, especially when the ratio of missing pixels was high.
In the reconstructed images in the last three rows for the non-smooth methods (i.e., IALM and HaLRTC), it was difficult to recognize the shapes of objects.
In the reconstructed images in the second to last row, the smooth matrix completion method (i.e., LTVNN) was not able to reconstruct the woman's facial features, but it is not difficult to recognize the woman's facial features in the second to last row where STDC, FBCP-MP, and SPC were applied.
For the last row with a 99\% missing ratio, the shape of ``Lena'' could be recognized only for FBCP-MP and the proposed SPC-QV, but the performance of SPC-QV was better than for FBCP-MP.

\begin{figure}[t]
\centering
\includegraphics[width = \textwidthh]{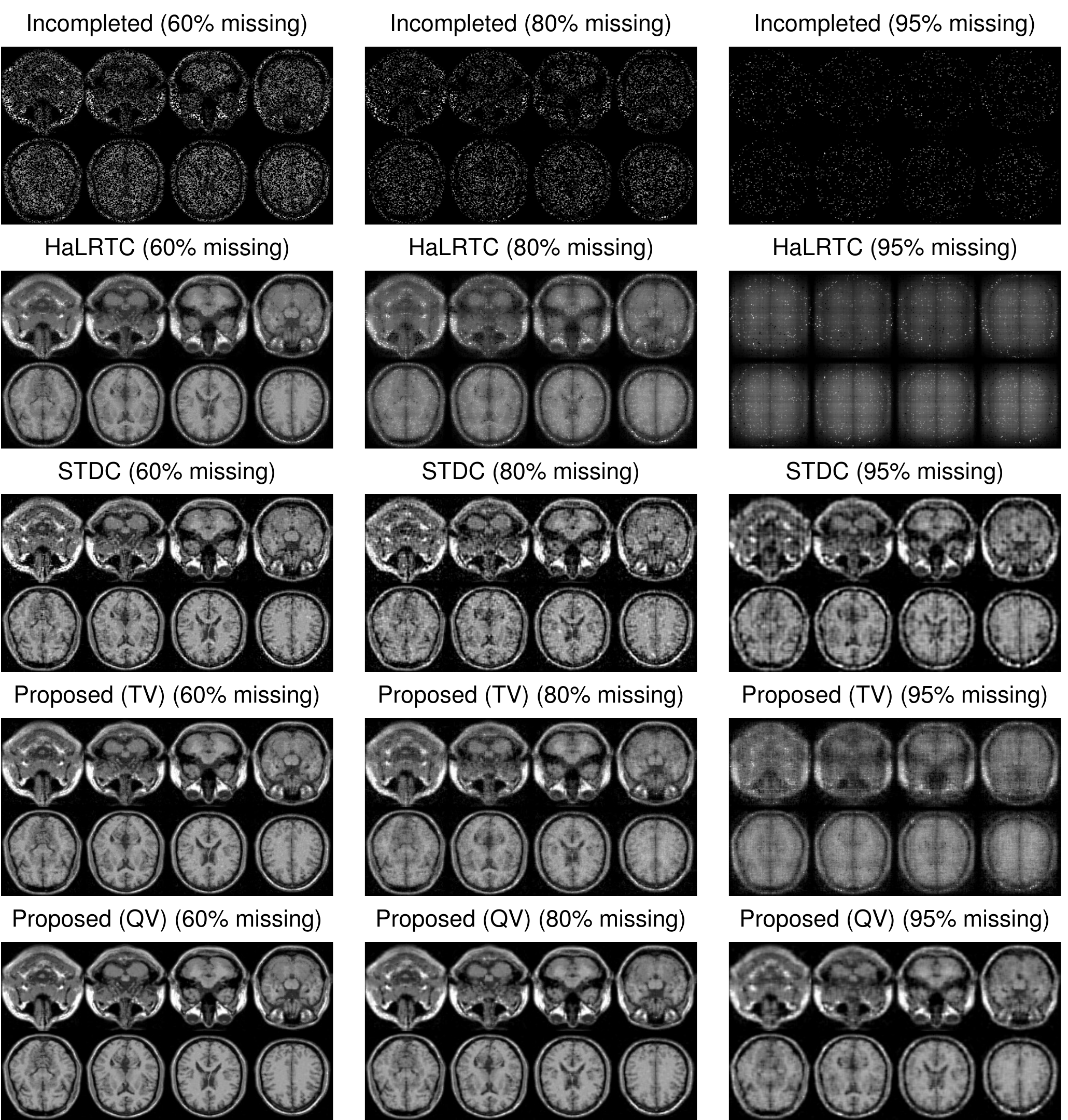}
\caption{Results of tensor completion for MRI data by using HaLRTC, STDC, and the proposed SPC -TV and SPC -QV methods for various missing rates.}\label{fig:MRI}
\end{figure}

\begin{figure}[t]
\centering
\includegraphics[width = \textwidthh]{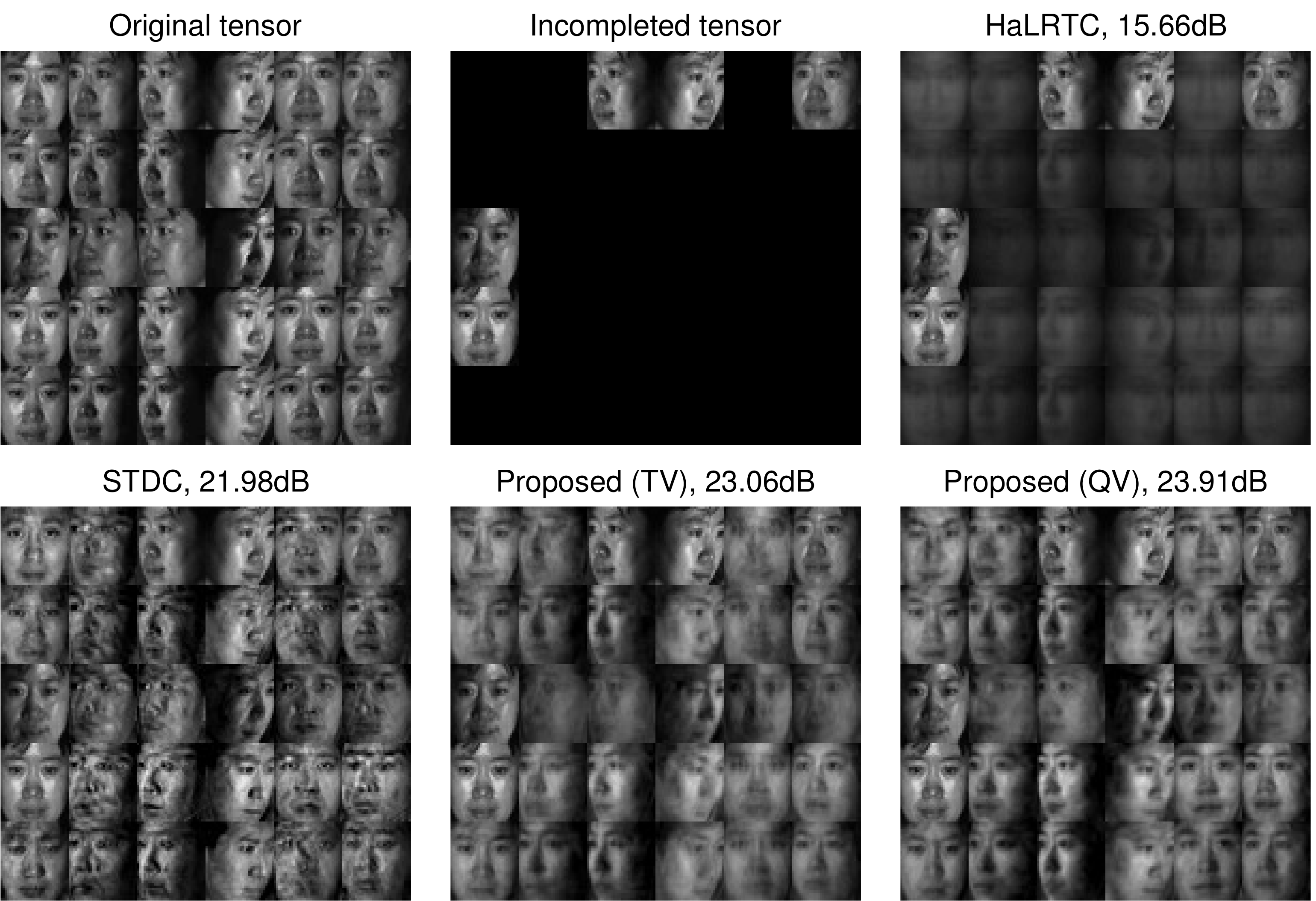}
\caption{Results of tensor completion for the CMU face dataset by using HaLRTC, STDC, and the proposed SPC -TV and SPC -QV methods. Values of SDR [dB] are described in each subfigure. Considerable improvements in performance were achieved.}\label{fig:CMU}
\end{figure}

\subsection{MRI Image Completion}
Fig.~\ref{fig:MRI} shows the results for several slices of MRI 3D-images (of size $109 \times 91 \times 91$), with 60\%-95\% of voxels missing, obtained by HaLRTC, STDC, and SPC with TV and QV smoothing.
Since the MRI image is smooth in all three dimensions (modes), the hyper-parameters of the SPC algorithm were set as $\bm\rho=[0.01, 0.01, 0.01]$ for TV smoothing, and $\bm\rho = [0.5, 0.5, 0.5]$ for QV smoothing.
The hyper-parameters of other methods were tuned optimally and manually.
Table~\ref{tab:MRI} shows the performances (SDR) of the individual algorithms.
From Fig.~\ref{fig:MRI} and Table~\ref{tab:MRI}, we can observe that the SPC algorithm succeeded in completing the incomplete MRI 3D-images, even with a 90\%-95\% missing ratio of voxels, and also significantly outperformed the other methods, with respect to the SDR.

\subsection{4th-order Tensor Completion Using a CMU Faces Dataset}
Next, we applied the three completion methods to a facial image database, provided by Carnegie Mellon University (CMU), called ``CMU faces'' \cite{sim2003cmu}.
The CMU dataset consists of 6930 facial images of size 32 $\times$ 32 pixels.
The 6930 facial images represent 30 individual people, 11 poses, and 21 kinds of illumination (i.e., 6930 = 30 $\times$ 11 $\times$ 21).
Thus, the size of the CMU dataset is ($30 \times 11 \times 21 \times 1024$).
Since each facial image is generally smooth in this dataset, but the neighboring people, poses, and illuminations are not always similar, we enforced the smoothness constraint in only the final fourth dimension.
Each facial image is described as a $1024$-dimensional vector, which is a vectorized from a $32 \times 32$-gray-scaled-image.
Therefore, in this experiment we employed a smoothness constraint matrix that applies a combination of vertical and horizontal smoothness: 
\begin{align}
  \bm L^{(n)} = \begin{pmatrix}\bm L_v^{(n)} \\ \bm L_h^{(n)} \end{pmatrix},
\end{align}
where $\bm L_v^{(n)}$ and $\bm L_h^{(n)}$ are the vertical and horizontal smoothness matrices, respectively.
The smoothness parameters of the SPC were set as $\bm\rho=[0, 0, 0, 0.01]$ for TV smoothing and $\bm\rho=[0, 0, 0, 0.1]$ for QV smoothing.
Fig.~\ref{fig:CMU} shows a section of the original and incomplete CMU faces, with 80\% of faces missing, and its completed results obtained by the HaLRTC, STDC, and SPC algorithms.
In this case, HaLRTC failed to complete the faces, while STDC provided several broken faces. However SPC achieved excellent results.

\begin{table}[t]
\caption{Signal to distortion ratio of MRI completion}\label{tab:MRI}
\centering
\begin{tabular}{ l || c c c c c }
methods       &  60\%  & 70\%  & 80\%  & 90\%  & 95\%  \\ \hline
HaLRTC        &  14.59 & 12.40 & 9.95  & 6.93  & 4.99  \\
STDC          &  14.97 & 13.78 & 13.28 & 12.81 & 11.46 \\ 
Proposed (TV) &  18.13 & 16.23 & 14.16 & 11.20 &  8.36 \\
Proposed (QV) &\bf  20.65 &\bf 19.09 &\bf 17.29 &\bf 14.73 &\bf 12.79 \\
\end{tabular}
\end{table}

\section{Discussions}\label{sec:disc}
\subsection{Smooth PARAFAC Decomposition Model}\label{sec:novelty}
In Section~\ref{sec:existing}, we introduced three methods for ``low-rank'' and ``smooth'' matrix/tensor completion: LTVNN \cite{han2014linear}, STDC \cite{chen2014simultaneous}, and FBCP-MP \cite{zhao2015bayesian}.
We will now expand on the differences between these models and our proposed model.

First, our smooth tensor decomposition model is based on enforcing the ``local similarity between neighbor elements in each component,'' i.e., the vectors of the factor (component) matrices.
In contrast, LTVNN \cite{han2014linear} imposes the smoothness constraint on the ``output matrix itself,'' but not on the ``components.''
Furthermore, the smoothness of both STDC \cite{chen2014simultaneous} and FBCP-MP \cite{zhao2015bayesian} is designed to enforce the similarity (or smoothness) between ``individual components.''
Thus our method is quite different from above methods \cite{han2014linear,chen2014simultaneous,zhao2015bayesian}.
In \cite{sorber2015selected}, a generalized framework for tensor factorization with missing entries is proposed and smooth factor constraint is included in this framework.
However, it is considering fixed-rank optimization problem, and does not propose model selection method for $R$.

Second, our method is enforcing the different levels of smoothness for different components adaptively.
Note that the smoothness constraint term in our objective function \eqref{criterion:SmCPTC} is multiplied by $g_r^2$.
Usually, the distribution of $|g_r|$ of the PARAFAC decomposition is similar to the distribution of the ``singular values'' of the singular value decomposition, which may decrease exponentially from largest to smallest especially for smooth image data.
Thus, our model can be interpreted as constructing a smooth image by building up the various levels of smooth components, beginning with the smoothest one.
It should be emphasized that the adaptive smoothness in our model is a key-technique to perform large improvements from state-of-the-art methods for image completion.


\subsection{TV Smoothing and QV Smoothing}\label{sec:dics:TVQV}
In this paper, we consider two types of smoothness constraints: TV and QV smoothing.
In image recovery problems, such as denoising and restoration, it is generally considered that the TV constraint is better than QV.
On the other hand, our result implies that the QV constraint performs better than TV for the image completion problem.
This is a new and important discovery for the image completion research community, because many studies still consider the TV constraint for completion problems \cite{dai2009physics,han2014linear}.

The advantages of QV for the completion problem can be summarized as follows.
If we consider the minimization of the TV/QV term of a signal $\bm z = [0, \text{NaN}, 2, \text{NaN}, 0]^T$, the respective solutions are $\bm z_{TV} = [0, a, 2, b, 0]^T$, where $a,b \in [0,2]$ and $\bm z_{QV} = [0, 1, 2, 1, 0]^T$.
Since the QV term has a strong convexity compared with the TV term, the uniqueness of $\bm z_{QV}$ can be enforced.
Furthermore, $z_{TV}$ has strong possibility of performing bumps in a component such as $[0,0,2,0,0]^T$, and including $[0,2,2,2,0]^T$ in the solution set.
In cases with high missing ratio, such as 90\% or 95\%, such bumps will be performed very often by TV minimization.


\subsection{Rank Increasing Approach}\label{sec:disc:model_selection}
In this section, we discuss the some advantages of rank increasing model selection which is employed in our method.

First, rank-increasing approach is more suitable than rank-decreasing approach for our smooth PD model.
If there is no smoothness constraint, then the weak upper-bound of tensor rank is $\min_k (\prod_{n \neq k} I_n)$ for a general $N$-th order tensor $\ten{X} \in \bbR{I_1 \times I_2 \times \cdots \times I_N}$, because any $\ten{X}$ can be factorized by only one dense matrix $\bm U^{(k)} = \bm X_{(k)}$ and a unit matrix constructed as $\bm I = \bm U^{(1)} \odot \bm U^{(2)} \odot ... \odot \bm U^{(k-1)} \odot \bm U^{k+1} \odot ... \odot \bm U^{(N)} \in \bbR{\prod_{n \neq k} I_n \times \prod_{n \neq k} I_n} $.
However, such a unit matrix is not suitable for a smooth constraint, and it is difficult to estimate the upper-bound of the tensor rank, because it depends on the imposed level of smoothness.
In contrast, the rank-increasing approach can easily be applied to various problems in practice since it does not require prior information for the tensor rank.

Second, the rank-increasing scheme is useful for the initialization.
In general, even if we know the exact minimum (canonical) tensor rank of the smooth PARAFAC/Tucker decomposition for a completion problem, many local minima may exist for the optimization problem.
Considering several local optimal solutions for the problem, then the results of optimization methods will depend on the initialization.
Fig.~\ref{fig:optimization} shows the concept of the optimization process for the SPC algorithm.
The algorithm starts from a rank-one tensor ($R=1$), the initialization of which has no critical meaning at this moment.
The initialization of the rank-$R$ tensor factorization is given by the rank-$(R-1)$ tensor factorization, which gives a better initialization for the completion problem, because this initialization arises from the lower-rank tensor space.
When the algorithm is finally stopped, the SPC algorithm is able to find a good solution that is close to the lower-rank tensor space.

In recent years, rank-increasing approach is a relatively hot topic in matrix and tensor factorization models \cite{mishra2013low,dolgov2013alternating,tan2014riemannian,uschmajew2015greedy,yokota2015a}.
The main statement in these papers is that the rank-increasing (greedy) approach is more efficient for large scale and ill-posed setting than nuclear-norm based scheme.
Our method is an additional positive result in this topic.

\subsection{Convergence Property}
Unfortunately, unconstrained low-rank tensor approximation problem for higher ranks can be ill-posed or ill-conditioned in many cases \cite{de2008tensor}.
However, some other positive results have been also established in recent years:
Considering the unconstrained problem for low-rank approximation without missing values, a rank-one approximation by the ALS algorithm achieves a local convergence \cite{wang2014global,zhang2001rank}.
PARAFAC decomposition with smooth spline components preserves the local convergence of the ALS algorithm \cite{reis2002parafac}.
The local convergence of the rank-one tensor approximation with penalized smoothing is proven in \cite{allen2013multi}.
As the latest result, rank-one ALS with regularization can obtain global convergence via Lojasiewicz inequality \cite{uschmajew2015a}.
Please note that ALS and HALS are completely equivalent for rank-one tensor approximation $R=1$.

For the case of $R \geq 2$, some results of block coordinate descent (BCD) scheme can be applied for HALS.
In \cite{xu2013block}, local convergence of BCD scheme consisting of multiple convex sub-optimizations is obtained for non-negative tensor factorization model.
In \cite{karlsson2014parallel}, a regularized PD based tensor completion algorithm is proposed and its local convergence is obtained.
These results can be applied for our optimization problem when $p=2$ except unit-norm constraints.

Unfortunately, theoretical result about convergence of our algorithm is not obtained.
The difficulty of analysis for our model may be caused by ``unit-norm constraint.''
This problem will be addressed in our future works.


\section{Conclusions}\label{sec:conclusions}
In this paper, we proposed a new low-rank smooth PARAFAC decomposition method for tensor completion problems.
Our approach and algorithms are quite different from existing methods. Instead of setting the upper bound of the expected rank of the tensor, our algorithm increases the number of components gradually until the optimal rank is achieved.
We considered two types of smoothness constraint in our SPC method: total variation (TV) and quadratic variation (QV). Our method was shown to outperform the state-of-the-art algorithms, in particular HaLRTC, STDC, and FBCP-MP.

\section*{Acknowledgment}
The authors would like to thank Chao Li for inspiring discussions.
This work was supported by Japan Society for the Promotion of Science (JSPS) KAKENHI Grant Number 15K16067.

\end{document}